\RequirePackage{rotating}
\PassOptionsToPackage{dvipsnames}{xcolor}
\documentclass[acmsmall]{acmart}

\AtBeginDocument{%
	}

\setcopyright{acmcopyright}
\copyrightyear{2018}
\acmYear{2018}
\acmDOI{XXXXXXX.XXXXXXX}

\acmJournal{JACM}
\acmVolume{37}
\acmNumber{4}
\acmArticle{111}
\acmMonth{8}

\usepackage[english]{babel}
\usepackage[utf8]{inputenc}
\usepackage{amsmath}
\usepackage{hyperref}
\addto\extrasenglish{  
	\def\sectionautorefname{Section}
	  
}
\usepackage{booktabs}
\usepackage{mathtools}
\usepackage[nohyperlinks,printonlyused,withpage]{acronym}
\usepackage{tikz} 
\usepackage{longtable}
\usepackage{enumitem}
\usepackage{siunitx}
\usepackage{comment}
\usepackage{subcaption}
\usepackage{tabularx} 
\usetikzlibrary{decorations,matrix,shapes,arrows,arrows.meta,positioning}
\usepackage{varwidth}
\usepackage{wrapfig}
\usepackage[title,toc]{appendix}
\usepackage{bbm}

\newcommand{\ALL}[1]{\textcolor{magenta}{all: #1}}
\newcommand{\CN}[1]{\textcolor{red}{CN: #1}}
\newcommand{\LW}[1]{\textcolor{orange}{LW: #1}}
\newcommand{\TK}[1]{\textcolor{blue}{TK: #1}}
\newcommand{\LP}[1]{\textcolor{green}{LP: #1}}
\newcommand{\RG}[1]{\textcolor{teal}{RG: #1}}

\newtheorem{remark}{Remark}
\newtheorem{example}{Example}
\newtheorem{definition}{Definition}

\newcommand*{\Autoref}[1]{%
	\begingroup
	\def\sectionautorefname{Section}%
	\autoref{#1}%
	\endgroup%
}

\DeclareMathOperator{\agg}{agg}
\DeclareMathOperator{\btn}{BTN}
\DeclareMathOperator{\stn}{STN}
\DeclareMathOperator{\btndt}{BTN_{DT}}
\DeclareMathOperator{\stndt}{STN_{DT}}
\DeclareMathOperator{\alongreqdt}{a_{long,req,DT}}
\DeclareMathOperator{\alatreqdt}{a_{lat,req,DT}}
\DeclareMathOperator{\alongmin}{a_{long,min,DT}}
\DeclareMathOperator{\alatmin}{a_{lat,min,DT}}
\DeclareMathOperator{\Ima}{Im}
\DeclareMathOperator{\CR}{CR}
\DeclareMathOperator{\SP}{SP}
\DeclareMathOperator{\PA}{\mathit{PA}}
\DeclareMathOperator{\pa}{\mathit{pa}}
\DeclareMathOperator{\CP}{\mathit{cp}}
\DeclareMathOperator{\nCP}{\neg{\mathit{cp}}}
\DeclareMathOperator{\out}{Out}
\DeclareMathOperator{\DoOp}{do}
\newcommand{\Do}[1]{\DoOp({#1})}

\begin{document}
	
\title[Grasping Causal Explanations of Criticality]{Grasping Causality for the Explanation of Criticality for Automated Driving}


\ccsdesc[500]{Mathematics of computing~Causal networks}
\ccsdesc[500]{Computing methodologies~Causal reasoning and diagnostics}
\ccsdesc[500]{Computing methodologies~Modeling methodologies}

\author{Tjark Koopmann}
\orcid{0000-0002-4865-3911}
\email{tjark.koopmann@dlr.de}
\affiliation{
	\institution{German Aerospace Center (DLR) e.V.}
	\department{Institute of Systems Engineering for Future Mobility} 
	\city{Oldenburg} 
	\country{Germany}
}

\author{Christian Neurohr}
\orcid{0000-0001-8847-5147}
\email{christian.neurohr@dlr.de}
\affiliation{
	\institution{German Aerospace Center (DLR) e.V.}
	\department{Institute of Systems Engineering for Future Mobility} 
	\city{Oldenburg} 
	\country{Germany}
}

\author{Lina Putze}
\orcid{0000-0002-7443-1191}
\email{lina.putze@dlr.de}
\affiliation{
	\institution{German Aerospace Center (DLR) e.V.}
	\department{Institute of Systems Engineering for Future Mobility} 
	\city{Oldenburg} 
	\country{Germany}
}

\author{Lukas Westhofen}
\orcid{0000-0003-1065-4182}
\email{lukas.westhofen@dlr.de}
\affiliation{
	\institution{German Aerospace Center (DLR) e.V.}
	\department{Institute of Systems Engineering for Future Mobility} 
	\city{Oldenburg} 
	\country{Germany}
}

\author{Roman Gansch}
\affiliation{
	\institution{Robert Bosch GmbH}
	\city{Renningen} 
	\country{Germany}
}
\email{roman.gansch@de.bosch.com}

\author{Ahmad Adee}
\affiliation{
	\institution{Technical University of Kaiserslautern}
	\city{Kaiserslautern}
	\country{Germany}
}
\email{adee4102@gmail.com}

\renewcommand{\shortauthors}{Koopmann et al.}

\begin{abstract}
	The verification and validation of automated driving systems at SAE levels 4 and 5 is a multi-faceted challenge for which classical statistical considerations become infeasible.
	For this, contemporary approaches suggest a decomposition into scenario classes combined with statistical analysis thereof regarding the emergence of criticality. Unfortunately, these associational approaches may yield spurious inferences, or worse, fail to recognize the causalities leading to critical scenarios, which are, in turn, prerequisite for the development and safeguarding of automated driving systems.
	As to incorporate causal knowledge within these processes, this work introduces a formalization of causal queries whose answers facilitate a causal understanding of safety-relevant influencing factors for automated driving.
	This formalized causal knowledge can be used to specify and implement abstract safety principles that provably reduce the criticality associated with these influencing factors.
	Based on Judea Pearl's causal theory, we define a causal relation as a causal structure together with a context, both related to a domain ontology, where the focus lies on modeling the effect of such influencing factors on criticality as measured by a suitable metric.
	As to assess modeling quality, we suggest various quantities and evaluate them on a small example.
	Our main example is a causal relation for the well-known influencing factor of a reduced coefficient of friction and its effect on longitudinal and lateral acceleration as required by the driving task.
	As availability and quality of data are imperative for validly estimating answers to the causal queries, we also discuss requirements on real-world and synthetic data acquisition.
	We thereby contribute to establishing causal considerations at the heart of the safety processes that are urgently needed as to ensure the safe operation of automated driving systems.
\end{abstract}
\vspace{-10em}
\maketitle

\section{Introduction}
\label{sec:introduction}

Traffic safety research is inherently tied to the investigation of causal questions such as 'which chain of events led to the accident?', 'could the accident have been prevented by a strong steering maneuver?' or 'are strict speed limits effective at increasing general safety in traffic?'.
In order to prevent accidents, besides the practical aspects of operating a car, the theoretical part of teaching humans to drive reasonably safe usually includes the numerous abstract classes of danger ('a wet road surface') combined with a causal explanation as to why an instance of this danger could lead to a critical situation ('less available tire-to-road friction implies decreased maneuverability'). Moreover, this abstract danger is accompanied by some safety principle (SP) to mitigate the potential risk ('if road surface is wet, drive at reduced speed').
The human brain excels at identifying and mitigating instances of such abstract dangers in the open context of the traffic world due to a causal understanding derived from prior experiences (e.g.\ knowledge obtained from driving lessons or parents) leading to solid predictions even for unseen circumstances.
Moreover, humans intuitively perform counterfactual reasoning after traffic incidents (or any incident) to \emph{learn} from those experiences and hence to avoid such incidents in the future.

Automated driving systems (ADSs) at SAE levels 4 and 5 \cite{sae2021definitions} are complex systems which are expected to safely navigate in open contexts in general, and specifically through all instances of the same abstract classes of dangers, just as humans do.
Without obtaining a formalized understanding of the underlying causalities, it is hardly possible to transfer this knowledge to ADSs.
However, in order to adapt the human learning process for automated driving, a rigorous process of identification and formalization of the knowledge to be learned is necessary. 
Recent work proposes a methodical \emph{criticality analysis} for ADSs \cite{neucrit21}, aiming to systematically identify influencing factors associated with increased criticality in traffic and analyze the underlying causal relations. 
In the work at hand,
we combine the framework of \emph{causal theory} \cite{pearl2009causal} with the criticality analysis as a novel approach to tackle the problem of modeling and analyzing the causal relations of such \emph{criticality phenomena} for ADSs with formal rigor.
Explaining these relations of contextual factors with increased criticality by a causal model with formal semantics enables the derivation and implementation of SPs that can be quantitatively proven to mitigate criticality. 
Finally, such evidence on criticality mitigation can be generically leveraged in a quantitative safety argumentation.

The contributions of this work may be summarized as providing
\begin{itemize}
	\item a formalization of causal queries within the criticality analysis,
	\item the application of causal theory for the formal modeling and analysis of causal relations,
	\item the introduction of quantities to evaluate the modeling quality of such causal relations, and
	\item siginificant modeling efforts towards the \textsc{reduced coefficient of friction}'s causal relation.
\end{itemize}
These contributions differ from contemporary approaches for understanding causality in the context of vehicle development, e.g.\ fault tree analysis, in that they allow to cover the vast modeling complexity as required by safety considerations for automated driving at SAE levels 4 and 5 while providing a formal basis grounded in the theories of random variables and directed acyclic graphs.  


The following \autoref{sec:preliminaries} introduces the methodical foundations of the criticality analysis, as introduced in \cite{neucrit21}, the required concepts from causal theory \cite{pearl2009causal} and discusses how causal effects can be estimated from data as well as other relevant related work. 
In \autoref{sec:causal_relations} we explore how causal theory can be applied to causal relations of criticality phenomena by representing them as causal structures. We discuss requirements and modeling principles for causal structures that lead to a rigorous definition of a causal relation and its plausibilization.
As to assess the modeling quality of causal relations, we introduce various indicator quantities and evaluate them on a small example.
\Autoref{sec:example} provides a detailed exemplification of the proposed modeling approach for the criticality phenomenon \textsc{reduced coefficient of friction}. 
Further, we discuss in \autoref{sec:elicitation} how causal relations constructed in this way relate to requirement eliciation for data acquisition, for real world data and in simulation environments, followed by ideas for future work in \autoref{sec:conclusion}.

\section{Preliminaries}
\label{sec:preliminaries}

In this section, we briefly present the foundations and preliminary work for the to-be-combined aspects of this work, namely criticality analysis and causal theory.

\subsection{Introduction to the Criticality Analysis for Automated Driving Systems}
\label{subsec:intro_crit_ana}

Any complex system that operates within highly complex, open and unpredictable contexts needs to undergo a rigorous safety procedure before deployment \cite{poddey2019validation}. 
For ADSs at SAE levels 4 and 5, current scientific and industrial advances as e.g. driven by the PEGASUS family projects VVM and SET Level\footnote{\url{https://vvm-projekt.de/en}, \url{https://setlevel.de/en}} suggest an iterative scenario-based verification and validation process \cite{neurfund20}.
For this, a suitable first step can be a \emph{criticality analysis}, where the operational domain (OD) is analyzed, structured, decomposed and understood w.r.t. potentially safety-relevant influencing factors \cite{neucrit21}. 

Criticality can be roughly understood as the combined risk of all actors within a traffic situation or scenario \cite[Definition 1]{neucrit21}. 
The criticality analysis aims to identify factors within the open context that are associated with increased criticality, called \emph{criticality phenomena}, to understand the underlying causalities and to derive generic principles preventing their occurrence or mitigating their effects. 
Such SPs aim to reduce the causal effect of a criticality phenomenon (CP) on the measurable aspects of criticality\footnote{ISO 26262 defines 'safety mechanisms', a related albeit narrower concept for fault corrections of functional components \cite{iso26262}.}, enabling ADS designers to safeguard the product. 
The main artifacts are a) criticality phenomena b) their causal relations and c) safety principles. 

As an exemplary CP, which will also serve as a running example throughout this work, consider the \textsc{reduced coefficient of friction} between the road and the tires of an ADS-operated vehicle, referred to as \emph{ego} in the following.
The criticality analysis collects a finite and managable list of such factors in the OD.
Their explanations can then become the basis for generic SPs -- e.g. 'drive carefully during rain and freezing temperatures' --, mechanisms also taught to humans during driving school. 
There, such advice rests upon the causal understanding of the driving teacher, that, for example, the combination of rain and freezing temperatures can lead to ice on the road.
This is often causal for a reduced coefficient of friction, which in turn can increase the probability of experiencing unstable driving dynamics. 

If causalities are understood during the design process of an ADS and specific versions of generic SPs are implemented in the final product, the machine inherits a causal understanding of traffic that is prerequisite to homologation. 
Using purely associative implementations, such as machine learning algorithms, is prone to inefficient and unsafe behavior in the field. 
It is only by causal understanding that a traffic context becomes predictable and therefore safely managable. 

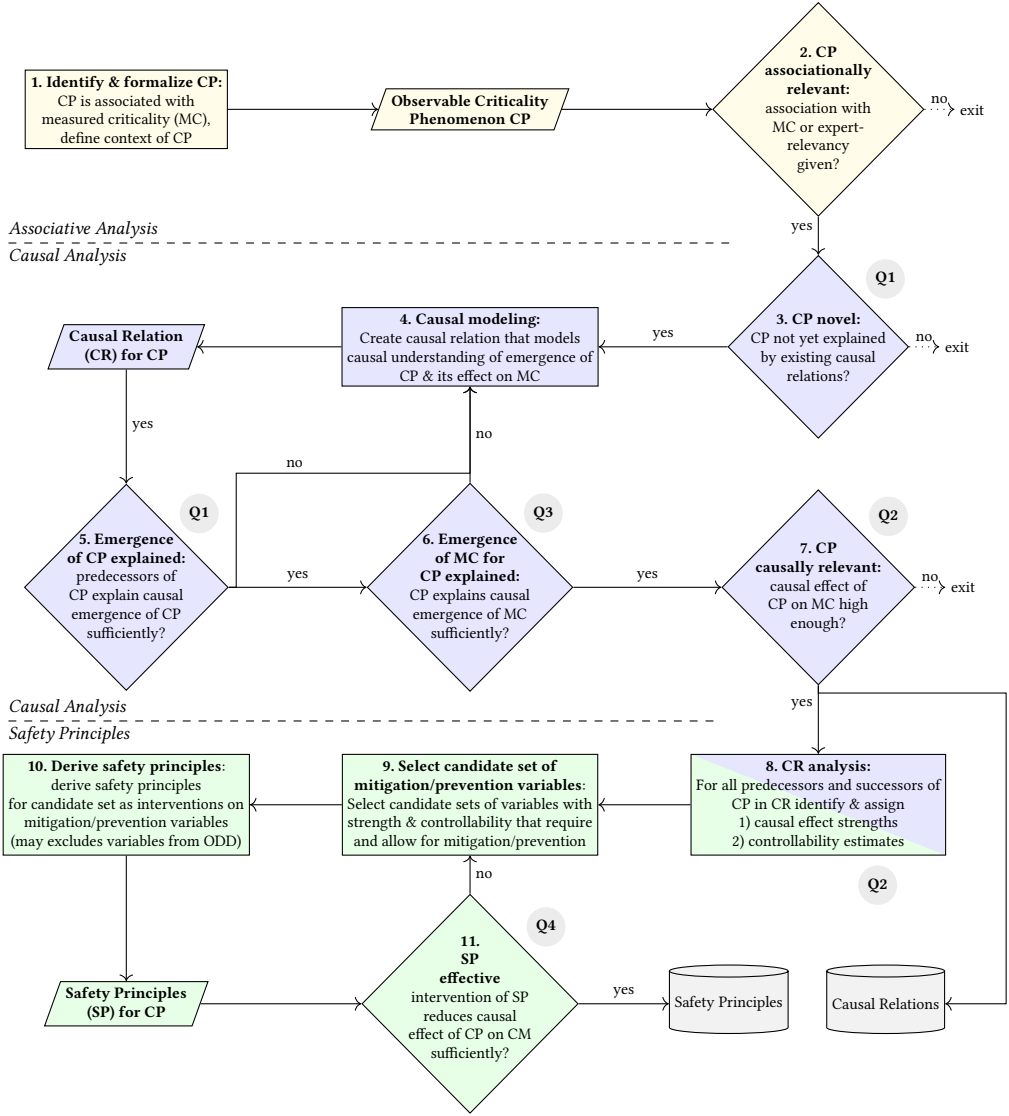
\begin{figure*}[htb!]
	\centering
\tiny
\tikzset{
	diagonal fill/.style 2 args={fill=#2, path picture={
			\fill[color=#1, sharp corners] (path picture bounding box.north west) -|
			(path picture bounding box.south east) -- cycle;}}}
\begin{tikzpicture}
	\node[draw, fill=yellow!10, align=center]		(1)	{\textbf{1. Identify \& formalize CP:}\\CP is associated with\\measured  criticality (MC),\\define context of CP};
	
	\node[draw, align=center, fill=yellow!10, align=center, right=2cm of 1, trapezium, trapezium left angle=70, trapezium right angle=110]	(2)	{\textbf{Observable Criticality}\\\textbf{Phenomenon CP}};
	
	\node[draw, diamond, fill=yellow!10, align=center, right=2cm of 2, inner sep=-2pt]	(3)	{\textbf{2. CP}\\\textbf{associationally}\\\textbf{relevant:}\\association with\\MC or expert-\\relevancy\\given?};
	\node[right=0.4cm of 3, align=center] (3ca) {exit};
	
	\node[draw, diamond, fill=blue!10, align=center, below=0.5cm of 3, inner sep=-2pt]	(4)	{\textbf{3. CP novel:}\\CP not yet explained\\by existing causal\\relations?};
	\node[right=0.4cm of 4, align=center] (4ab) {exit};
	\node[circle, fill=gray!15, above right=0.1cm and 0.1 cm of 4, align=center] (q1a) {\textbf{Q1}};
	
	\node[draw, fill=blue!10, align=center]	(5)	at (2 |- 4)	{\textbf{4. Causal modeling:}\\Create causal relation that models\\causal understanding of emergence of\\CP \& its effect on MC};
	
	\node[draw, align=center, fill=blue!10, align=center, trapezium, trapezium left angle=70, trapezium right angle=110]	(6)	at (1 |- 4)	{\textbf{Causal Relation}\\\textbf{(CR) for CP}};
	
	\node[draw, diamond, fill=blue!10, align=center, below=1.5cm of 6, inner sep=-2pt]	(7)	{\textbf{5. Emergence}\\\textbf{of CP explained:}\\predecessors of\\CP explain causal\\emergence of CP\\sufficiently?};
	\node[circle, fill=gray!15, above right=0.1cm and 0.1 cm of 7, align=center] (q2a) {\textbf{Q1}};
	
	\node[draw, diamond, fill=blue!10, align=center, inner sep=-2pt]	(8)	at (7 -| 5)	{\textbf{6. Emergence}\\\textbf{of MC for}\\\textbf{CP explained:}\\CP explains causal\\emergence of MC\\sufficiently?};
	\node[circle, fill=gray!15, above right=0.1cm and 0.1 cm of 8, align=center] (q1b) {\textbf{Q3}};
	
	\node[draw, diamond, fill=blue!10, align=center, inner sep=-2pt]	(9)	at (7 -| 4)	{\textbf{7. CP}\\\textbf{causally relevant:}\\causal effect of\\CP on MC high\\enough?};
	\node[right=0.4cm of 9, align=center] (9ca) {exit};
	\node[circle, fill=gray!15, above right=0.1cm and 0.1 cm of 9, align=center] (q3) {\textbf{Q2}};
	
	\node[draw, diagonal fill={blue!10}{green!10}, align=center, below=0.9cm of 9]	(10)	{\textbf{8. CR analysis:}\\For all predecessors and successors of\\CP in CR identify \& assign\\1) causal effect strengths\\2) controllability estimates};
	\node[circle, fill=gray!15, below right=0.2cm and -1.1 cm of 10, align=center] (q2b) {\textbf{Q2}};
	
	\node[draw, fill=green!10, align=center]	(11)	at (10 -| 8)	{\textbf{9. Select candidate set of}\\\textbf{mitigation/prevention variables}:\\Select candidate sets of variables with\\strength \& controllability that require\\and allow for mitigation/prevention};
	
	\node[draw, fill=green!10, align=center]	(12)	at (11 -| 7)	{\textbf{10. Derive safety principles}:\\derive safety principles\\for candidate set as interventions on\\mitigation/prevention variables\\(may excludes variables from ODD)};
	
	\node[draw, diamond, fill=green!10, align=center, inner sep=-2pt, below=0.5cm of 11]	(14)	{\\[-1cm]\textbf{11.}\\\textbf{SP}\\\textbf{effective}\\intervention of SP\\reduces causal\\effect of CP on CM\\sufficiently?};
	\node[circle, fill=gray!15, above right=0.1cm and 0.1 cm of 14, align=center] (q4) {\textbf{Q4}};
	
	\node[draw, align=center, fill=green!10, align=center, trapezium, trapezium left angle=70, trapezium right angle=110]	(13)	at	(14 -| 12)	{\textbf{Safety Principles}\\\textbf{(SP) for CP}};
	
	\node[right=1.2cm of 14, fill=gray!10, cylinder, draw, shape border rotate=90,shape aspect=0.1, minimum width=1.5cm, minimum height=0.9cm] (sps) {Safety Principles};
	
	\node[right=0.5cm of sps, cylinder, fill=gray!10, draw, shape border rotate=90,shape aspect=0.1, minimum width=1.5cm, minimum height=0.9cm] (crs) {Causal Relations};

	\node[below right=1.68cm and -2.6cm of 3.east] (div1r) {};
	\node[] (div1l)	at	(12.west |- div1r) {};
	\node[above=0.10cm of div1l, anchor=west] (div1) {\scriptsize\emph{Associative Analysis}};
	\node[below=0.10cm of div1l, anchor=west] (div1) {\scriptsize\emph{Causal Analysis}};
	
	\node[below right=1.68cm and -2.6cm of 9.east] (div2r) {};
	\node[] (div2l)	at	(12.west |- div2r) {};
	\node[above=0.10cm of div2l, anchor=west] (div1) {\scriptsize\emph{Causal Analysis}};
	\node[below=0.10cm of div2l, anchor=west] (div1) {\scriptsize\emph{Safety Principles}};

	\draw[->]	(1)		edge												(2);
	\draw[->]	(2)		edge												(3);
	\draw[->]	(3)		edge	node[left, yshift=0.1cm]	{yes}			(4);
	\draw[->]	(4)		edge	node[above]	{yes}							(5);
	\draw[->]	(5)		edge												(6);
	\draw[->]	(6)		edge	node[right]					{yes}			(7);
	\draw[->]	(7)		edge	node[above]					{yes}			(8);
	\draw[->]	(8)		edge	node[above]					{yes}			(9);
	\draw[->]	(8)		edge	node[right]					{no}			(5);
	\draw[->]	(9)		edge	node[left, yshift=0.2cm]	{yes}			(10);
	\draw[->]	(7.east) -| ++(0.1,1.5)  -|	node[above, very near start]	{no}	(5);
	\draw[->]	(9.south) -- ++(0,-0.1) -- ++(2.5,0) |- 					(crs);
	\draw[->]	(10)	edge												(11);
	\draw[->]	(11)	edge												(12);
	\draw[->]	(12)	edge												(13);
	\draw[->]	(13)	edge												(14);
	\draw[->]	(14)	edge			node[right]			{no}			(11);
	\draw[->]	(14)	edge			node[above]			{yes}			(sps);
	\draw[->]	(3)		edge[dotted]	node[above]			{no}			(3ca);
	\draw[->]	(4)		edge[dotted]	node[above]			{no}			(4ab);
	\draw[->]	(9)		edge[dotted]	node[above]			{no}			(9ca);
	
	\draw[]		(div1l)	edge[dashed]										(div1r);
	\draw[]		(div2l)	edge[dashed]										(div2r);
\end{tikzpicture}
	\Description{The methodical part of the criticality analysis for unveiling causalities behind criticality and deriving effective safety principles. Relevant steps are annotated with their respective causal queries \textbf{Q1} to \textbf{Q4}.}
	\caption{The methodical part of the criticality analysis for unveiling causalities behind criticality and deriving effective safety principles. Relevant steps are annotated with their respective causal queries \textbf{Q1} to \textbf{Q4}.}
	\label{fig:process}
\end{figure*}

This work closely examines the role of the mandatory causal analysis of criticality within an open and complex context. 
The concept paper of the criticality analysis provides a rather syntactical definition of a \emph{causal relation} as a \emph{directed, acyclic graph  $\mathit{CR} = (P, E)$ where $P$ is the set of nodes -- variables described by propositions on traffic scenarios -- and $E \subseteq P \times P$ is the set of edges -- causal links between the variables} \cite[Definition~7]{neucrit21}.
This work aims at sharpening this definition using Pearl's causal theory. 
From the point of view of the criticality analysis, we focus on the method branch, as introduced in \cite[Figure~4]{neucrit21}, and extend it to incorporate SPs, leading to the basic process sketched in \autoref{fig:process}.
Here, we start with a CP, ensure its observability and establish its associational relevance. 
Thereafter, the causal analysis starts by determining the novelty of the phenomenon. 
A core part is the creation of a causal relation, reflecting the understanding of the analyst about the emergence of the CP as well as its effect on measured criticality. 
This causal relation enables predictions and serves as a backbone for subsequent analyses and queries.

Firstly, the causal relation is checked for plausibility: both in its assumptions on the emergence as well as its measurable effect. 
Afterwards, the causal effect of the CP on measured criticality is analyzed.
If it is too low (e.g.\ due to a previously identified spurious association), we exit the analysis. 
If this is not the case, we are presented with a CP that is both warranted for mitigation or prevention as well as causally well understood. 
Hence, we analyze the causal relation w.r.t.\ the variables that are suitable candidates for mitigation or prevention, i.e.\ variables that are sufficient in causal effect and controllability at design- or runtime. 
Once candidate variables are identified, we derive SPs in the form of interventions. 
Finally, it has to be decided whether the causal effect is sufficient, otherwise the candidate variables need to be refined. 

This process intentionally relies on causal language. 
We condense the causal queries to the following four aspects and indicate which steps of \autoref{fig:process} require their evaluation:

\begin{itemize}
	\item[\textbf{Q1}] The explanation of a CP by a set of predecessors in a causal relation (\emph{steps 3, 5}) 
	\item[\textbf{Q2}] The causal effect of a CP on criticality as measured by a suitable metric (\emph{steps 7, 8}) 
	\item[\textbf{Q3}] The explanation of measured criticality by a CP (\emph{step 6}) 
	\item[\textbf{Q4}] The causal effect of SPs on reducing measured criticality (\emph{step 11})
\end{itemize}

A central aspect of this work, which is presented in \autoref{sec:causal_relations}, is the formalization of these queries using the framework of the do-calculus, as introduced in \autoref{subsec:intro_causal_models}.
In the remainder of this introduction, we will provide an intuitive understanding of the queries using the previously sketched example as well as motivate the need for their formalization and for causal inference.

Recall that we are investigating the CP \textsc{reduced coefficient of friction}, where we formalize the coefficient of friction $\mu$ being 'reduced' as $\mu \leq 0.4$. 
Suspecting that a \textsc{reduced coefficient of friction} negatively impacts a vehicle's maneuverability, we measure criticality by $\max(\stn, \btn)$, the maximum of the Steer resp. Brake Threat Number \cite[Section~5.2]{westhofen2022metrics}.

Assuming that the CP is novel (\textbf{Q1}), i.e.\ it is not sufficiently explained by some existing causal relation, e.g.\ for \textsc{slick road surface}, the analyst formalizes their understanding of the CP by causal assumptions on the emergence of a reduced road-tire friction and its impact on $\btn$ and $\stn$. 

We can now use this model to answer causal queries.
As a first step, it is imperative to validate the causal assumptions: 
does the model explain the presence of $\mu \leq 0.4$ within the population (\textbf{Q1}), e.g.\ by incorporating the influences of weather, temperature, vehicle dynamics and vehicle tires, appropriately?
If so, we subsequently ask if $\mu \leq 0.4$ explains the increase of $\max(\stn, \btn)$ to a satisfactory extent (\textbf{Q3}).
If those queries have been investigated and successfully compared to acceptable thresholds, it is essential to understand the strength of the effect of a \textsc{reduced coefficient of friction}, i.e.\ $\mu \leq 0.4$, on criticality as measured by $\max(\stn, \btn)$ (\textbf{Q2}). 
This gives us a notion of relevance -- based on the amplitude of the causal effect, we can decide whether to further examine the CP.



Assuming relevance of the CP, if the queries have been investigated and successfully compared to acceptable thresholds, SPs that reduce the impact of reduced road-tire friction can be derived from this causal model.
Causal theory delivers us with formal tools in guiding such a process. 
By examining the strength of the causal effect of predecessor and successor variables of the CP on the CP resp. measured criticality in the causal model (\textbf{Q2}), we are able to identify suitable candidates. 
In our example, we may decide that the set of mitigation/prevention variables $\{ \text{precipitation, temperature} \}$ has a large effect on $\mu$ and are controllable in the sense that it can be excluded from the ADS's operational design domain (ODD), i.e.\ the ADS shall not operate under those circumstances. 
A derived $\SP$ then defines value ranges for the identified variables. 
For example, the ADS shall not operate whenever precipitation $\geq 10$ mm/h and temperature $\leq 0$°C. 

$\SP$s target the CP either by reducing its probability of occurrence or by mitigating its effects downstream. 
Note that this differs from defining $\SP$s independently of CPs. 
Consider a $\SP$ that does not specifically target the causal relation for the coefficient of friction, but influences the $\btn$ and $\stn$ such as 'always hold a minimum distance of $5$m to other traffic participants'.
This $\SP$ reduces the $\btn$ and $\stn$ as well, but is not based on specific causal findings.
This is problematic as
\begin{enumerate}
	\item we have no guarantee about the generalizability of the SP to situations with similar causalities (e.g.\ swerving on icy roads without other traffic participants),
	\item in a safety case, there is hence no possibility to argument  that a certain abstract class of danger, i.e.\ a CP, is mitigated by implementing a SP (e.g.\ the safety argumentation may only use the evidence that $\btn$ and $\stn$ are reduced on average), and
	\item such unspecific SPs lead to driving functions that may perform inefficiently outside the abstract classes of dangers (e.g.\ holding unnecessary distance to parking vehicles). 
\end{enumerate}
This shows that SPs ought to be checked for effectiveness (\textbf{Q4}). 
In our example, we can analyze whether an intervention on precipitation $< 10$ mm/h or temperature $> 0$°C leads to a reduction
\begin{itemize}
	\item in the probability of occurrence of the CP \textsc{reduced coefficient of friction}, or
	\item to a reduction of the CP's effect on criticality as measured by $\max(\stn, \btn)$.
\end{itemize}
If so, this is one of various valid $\SP$s that can be used for the design and implementation of an ADS. 

Here, we have sketched how causal analysis constructively contributes to safeguarding ADSs, which will be addressed in-depth for the remainder of this work. In particular, we provide a rigorous theoretical foundation and a detailed example of a causal relation.

\subsection{Introduction to Causal Theory}
\label{subsec:intro_causal_models}

The systematic investigation of causal questions requires a formal expression of causal relationships.
Traditional concepts of probability theory can infer associations and estimate the probability of past and future events but they lack in representing causal relationships.
Pearl complements these concepts by introducing a framework for causal inference \cite{pearl_2009, pearl2009causal}, which combines the representation of causal relationships by graphical modeling and the analysis of causal queries based on Bayesian stochastics.
In this section we will give an overview of the main ideas and concepts of Pearl's \textit{causal theory} \cite{pearl_2009, pearl2009causal}. that we apply in the context of the criticality analysis in \autoref{sec:causal_relations}. 

\begin{figure}[htb!]
	\centering
	\begin{subfigure}{0.49\linewidth}
		\centering
		\includegraphics[width=0.85\linewidth]{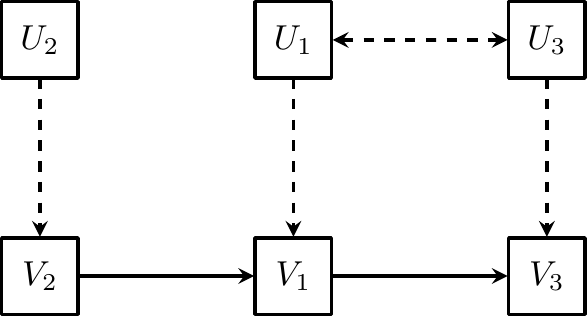}
		\caption{Example of a causal structure.}
		\label{fig:causal_structure_example}
	\end{subfigure}
	\hfill
	\begin{subfigure}{0.49\linewidth}
		\centering
		\includegraphics[width=0.85\linewidth]{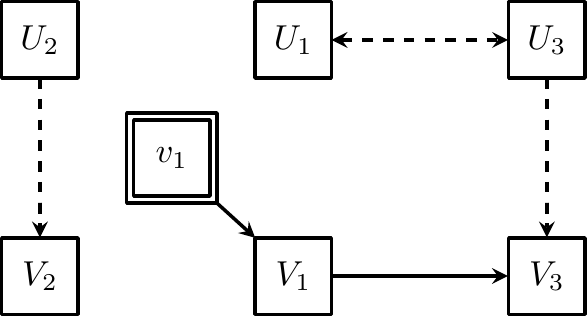}
		\caption{Causal structure post-intervention for $\Do{V_1 = v_1}$}
		\label{fig:post_intervention_example}
	\end{subfigure}
	\Description{Example of a causal structure pre- and post-intervention with dependent and independent error terms.}
	\caption{Example of a causal structure pre- and post-intervention with dependent and independent error terms.}
	\label{fig:causal_structure_both}
\end{figure}

Pearl describes the causal modeling as an 'identification game that scientists play against nature' \cite[p.43]{pearl_2009}. 
The main idea is that in nature on a sufficient level of detail  there exist deterministic functional relationships between different variables. These relationships can be represented in a graphical structure that the scientist tries to  identify. 
The so-called \emph{causal structure}, denoted by $\mathfrak{S}$, of a set of variables is defined as a directed acyclic graph (DAG) where the edges represent possible direct functional relationship between the nodes which represent the variables.
Any consecutive sequence of arrows is called a \emph{path}, regardless of the orientations of the arrows.
A distinction is made between endogenous variables $V$, which are determined by other variables in the graph, and exogenous variables $U$, also called \emph{error terms} or \emph{disturbances}, which represent background factors that are determined by factors outside of the causal structure. 
Those background factors might be correlated which is usually illustrated by dashed double arrows between the variables as shown in \autoref{fig:causal_structure_both}.   
We remark that an arrow in a causal structure models the possibility of a direct causal influence (or of a correlation in case of the dashed double arrows). 
Therefore, it is the absence of an arrow which provides the causal information. 
As an example, the missing arrow in \autoref{fig:causal_structure_example}
between $Y$ and $Z$ models the assumption that $Y$ has no direct influence on $Z$. However, there exists the possibility of an indirect influence along the path $Y\rightarrow X\rightarrow Z$. Moreover, the error term $U_Y$  is supposed to be independent of $U_X$  and $U_Z$ as illustrated by the absence of dashed double arrows.

A \emph{causal model} specifies a causal structure by defining how each variable is influenced by its parents.
More precisely, the model $\mathfrak{M} = (\{f_i\}_i, P_U)$ consists of a joint probability distribution $P_U$ of the exogenous variables $U_i$ and a function $f_i(pa_i, u_i)$ for each endogenous variable $V_i$, which determines the variable based on their endogenous parent nodes $\PA_i$  and the related exogenous error terms $U_i$. 
Additionally, for a causal model independence of the error terms is demanded. 
Hence, a causal model $\mathfrak{M}$  defines a joint probability distribution $P(\mathfrak{M})$ over the variables in the causal structure which reflects the conditions of dependency. 

The independence of the exogenous variables in a causal model combined with the acyclic shape of the causal structure ensures that  each variable is independent of all its non-descendants, given its parents -- a characteristic called \emph{Markov condition}. 
Pearl states that a model whose variables are explained in microscopic detail is assumed to satisfy the conditions of a Markovian model, but the aggregation of  variables and their corresponding probabilities on  higher abstraction levels might lead to cycles in the structure or dependencies of the error terms \cite[p.44]{pearl_2009}.
As the Markov condition indicates  whether a set of parent nodes is considered to be complete in the sense that all relevant immediate causes of a node are modeled it can serve as a criterion for the right level of abstraction.

Based on the theory of causal models the causal effect can be analyzed by the so-called \emph{do-operator}. 
Given a set of variables $X$, the operator $\Do{X=x}$, abbreviated as $\Do{x}$, simulates an atomic physical intervention. 
This is implemented by deleting the functions which define the variables in $X$ from the model and substituting $X=x$ in the other functions while keeping the rest of the model unchanged, cf.~\autoref{fig:post_intervention_example}. 
The respective post-interventional model is denoted by $\mathfrak{M}_x$. 
Therefore, the post-interventional distribution  of an outcome $Y$ can be described as the probability distribution that the model $\mathfrak{M}_x$ assigns to $Y$:
\begin{equation*}
	P_\mathfrak{M}(Y|\Do{X=x})=P_{\mathfrak{M}_x}(Y)\,.
\end{equation*}
In order to estimate the post-interventional distribution, conventional statistical approaches implement the intervention in an experimental setup, e.g.\ as a randomized control trial. 
However, such a direct manipulation may not always be feasible due to physical, ethical or economic reasons. 
The do-calculus provides the fundamentals for estimating the causal effect based on the pre-interventional distribution without additional experimental data. 
This characteristic is called \emph{identifiabilty}: 
A causal effect is identifiable from a graph if it can be computed uniquely from any positive probability of the observed variables.

Given a Markovian model the causal effect of a set of variables $X$ on a disjoint set $Y$ is identifiable if $X$, $Y$ and all parents $\PA_X$ of variables in $X$ are measurable \cite[Theorem 3.2.5]{pearl_2009}. 
Then, we can write
\begin{equation}\label{EQ Markov 1}
	P(Y=y | \Do{X=x}) = \sum_{\pa_x}P(Y=y|X=x, \PA_x=\pa_x)P(\PA_x=\pa_x)\,.
\end{equation}
A special case holds when all variables in a Markovian model $\mathfrak{M}$ are measured: 
As the Markovian property of $\mathfrak{M}$ is inherited to any interventional model $\mathfrak{M}_x$, the joint post-interventional distribution of all variables $X,  V_1,\dots, V_n$ regarding an intervention $\Do{X=x}$ can be expressed by
\begin{equation*}
	P(X=x, V_1=v_1, \dots V_n=v_n | \Do{X=x}) = \prod\limits_{i=1,\dots,n} P(V_i=v_i |\PA_i=\pa_i)|_{X=x}\,,                           
\end{equation*}
where $\PA_i$ describes the set of parent nodes of the variable $V_i$ \cite[Corollary 1]{pearl2009causal}.
The causal effect of $X=x$ on a variable $V_j$ can then be derived by marginalization, which means that we sum over all possible values  the variables  $V_1,\dots,V_{j-1},V_{j+1},\dots, V_n$ can take:
\begin{equation}\label{EQ Markov 2}
	P(V_j=v_j | \Do{X=x}) = \sum_{(v_1,\dots, v_{j-1},v_{j+1},\dots ,v_n)} \prod\limits_{i=1,\dots,n} P(V_i=v_i |\PA_i=\pa_i)|_{X=x}\,. 
\end{equation}
Both equations \eqref{EQ Markov 1} and \eqref{EQ Markov 2}  do not involve any assumptions regarding the specific functions $\{f_i\}_i$ or the distribution of the error terms $P_U$. 
The causal effect is just estimated based on the structure of the graph and the measured data.  
In addition, equation \eqref{EQ Markov 1} implies that measuring all parent nodes of $X$ is sufficient to estimate the causal effect.
However, there might be other sets of variables besides the parent nodes which are sufficient to ensure the identifiability of the causal effect. 
The \emph{back-door criterion} provides some conditions to determine whether a set of measurable variables is sufficient to estimate the causal effect.  
The main idea is to choose a set of variables that blocks all spurious associations between $X$ and $Y$ and that does not open any new spurious associations by controlling for a collider. 
In the language of causal models, a path $p$ between two variables $X$ and $Y$ is called  \emph{blocked} (or \emph{d-separated}) by a set of nodes $S$ if  either $p$ contains a chain $i\rightarrow m \rightarrow j$ or a fork $i\leftarrow m\rightarrow j$ such that the middle node $m$ is in $S$ or if $p$ contains an inverted fork $i\rightarrow m \leftarrow j$  such that that neither the collider $m$, nor its descendants are in $S$.

The back-door criterion states that a set $S$ of measurable variables is \textit{admissible for adjustment} if no element of $S$ is a descendant of $X$ and the elements of $S$ block all paths from $X$ to $Y$ that contain an arrow into $X$. 
For such $S$, the causal effect of $X$ on $Y$ is identifiable by \cite[Theorem 3.3.2]{pearl_2009}:
\begin{equation}
	\label{eq:causal_effect_adjustmentset}
	P(Y=y\mid \Do{X=x}) = \sum_s P(Y=y\mid X=x,S=s)P(S=s)\,.
\end{equation}
The criterion holds for graphs of any shape, even for semi-Markovian models, i.e.\ models whose error terms are not necessarily independent.
Based on the back-door criterion, adjustment sets can be computed \cite{Tian1998}, e.g.\ using pgmpy \cite{ankan_pgmpy_2015}, DAGitty \cite{textor_2011_DAGitty} and DoWhy \cite{sharma2020}. 
This work relies on pgmpy. 


\subsection{Relation to Randomized Controlled Trials}
\label{subsubsec:estimating_causal_effects}

In many scientific fields, e.g. biology, medicine and psychology, the study design of randomized controlled trials (RCTs) has regularly been used for the estimation of causal effects.
RCTs rely on experimental studies while addressing the problem of confounding by randomization which means that the effect of an intervention on an outcome is estimated by comparison of randomly assigned groups, e.g.\ 'treatment group' vs. 'placebo group'.
RCTs are designed to quantify a single cause-effect relationship in experimental set-ups, but they are constructed in a way that treats the underlying causal influences as a black box. 
In contrast, the criticality analysis seeks to comprehensively investigate the entire causal relation as outlined in \autoref{subsec:intro_crit_ana}.
Therefore, the  estimation of causal effects based on RCTs cannot compensate an insufficient modeling of the corresponding causalities.
Nevertheless, RCTs constitute an option for the quantitative assessment of the causal effect of a CP on criticality.
Further, they allow for quantifying single edges within the causal structure, proving the existence of the respective causal link.

However, the application of RCTs for the safeguarding of ADSs comes with several limitations.
First, we remark that a RCT solely makes statements about a single causal assumption. 
Related causal assumptions cannot be subsequently derived from it and require their own experimental set up.
This holds, in particular, for the investigation of influences of parent nodes as well as for suggested avoidance or safety principles required for the queries \textbf{Q2} and \textbf{Q4} and thus comes along with a huge effort.
Moreover, the complexity of influencing factors and their relations is hard to reproduce in other environments than real-world traffic. 
This means that  the results of RCTs performed on proving grounds or in simulations are typically not transferable without evidence that the underlying causal models in real-world traffic and the experimental set up of the simulation or of the proving ground are, at least, interventionally equivalent \cite[Definition 6.47]{peters2017}. 
A possible step towards transferable RCTs on proving grounds could be a hybrid reality approach where a Prototype-in-the-Loop (PiL) is combined with a Model/Software-in-the-Loop (MiL/SiL)  \cite{hallerbach2022simulation}.

Likewise, conducting RCTs in real-world traffic turns out to be difficult, as human lives can be affected by experiments conducted in traffic.
Consequently, the only experiments that are ethically justifiable are those expected to decrease the criticality, with reasonable certainty, or to have no influence. 
Moreover, an informed consent of the trial's participants is almost impossible to realize for experiments conducted in real-world traffic.
Furthermore, we remark that there are several physical and social restrictions that need to be considered when implementing interventions in real world, e.g. it is not justifiable to  reduce the coefficient of friction by artificially producing black ice.

A huge advantage of causal theory is the estimation of causal effects based on non-experimental data via the do-calculus. 
Although obtaining suitable non-experimental data can be challenging as well, its less restrictive than experimental set-ups.
The do-calculus allows performing multiple interventions on a single data basis so that comprehensive investigations regarding the influence of parent nodes or certain avoidance or SPs are quite natural to integrate. 
Furthermore, criteria as the Markov property of a model or the back-door criterion enable the handling of unmeasurable variables by reducing the data requirements to parent nodes or adjustment sets.

\subsection{Related Work}
\label{subsec:rel_work}

As to adequately relate this work to the literature, in this subsection, we elaborate on various pieces of research which lie at the intersection of automotive safety, causality and intelligent systems.

\subsubsection{Bayesian Networks in Safety Considerations}
Bayesian Networks \cite{marcot2019advances} are similar to causal models in that they are both probabilistic graphical models that represent a set of random variables and their conditional dependencies via a DAG. 
They have been considered for safety considerations, in particular for risk-analysis and reliability topics, in many domains \cite{weber2012overview}.
In his thesis \cite{rauschenbach2017probabilistische}, Rauschenbach laid out the probabilistic basis for modeling errors in complex systems using Bayesian networks.
Rauschenbach and Nuffer applied Bayesian networks in this context for a probabilistic Failure Mode and Effect Analysis (probFMEA) and for evaluation of functional safety metrics \cite{rauschenbach2019quantitative}.
Steps towards the application of Bayesian networks in the automotive domain have also been taken, e.g. to learn models for the probability distributions of initial scene configurations from real-world data \cite{wheeler2015initial}.
In recent work, Maier and Mottok present the tool \textit{BayesianSafety} that builds on Bayesian networks to combine environmental influences, machine learning and causal reasoning for automotive safety analyses \cite{maier_bayesiansafety_2022}.
However, causal effects cannot be inferred from the conditional probability distributions (CPDs) alone. For this, causal theory introduces the do-calculus.

\subsubsection{Causality in Automotive Safety}
The well-known automotive safety standards ISO 26262 \cite{iso26262} and ISO 21448 \cite{iso21448} already consider the problem of causality, in particular regarding the identification of factors that cause hazardous behavior.
For this, classical methods for hazard analysis and risk assessment are recommended, such as Fault Tree Analysis (FTA) \cite{vesely1981fault} or System Theoretic Process Analysis (STPA) \cite{ishimatsu2010modeling}.
In \cite{kramer2020identification,bode2019identifikation}, the authors present a safety analysis for ADS at SAE Level $\ge 3$ that relies on an extended FTA for causal analysis of hazards with a focus on triggering conditions in the sense of ISO 21448.
While these approaches focus on identification of hazards for a concrete item, the criticality analysis focuses on abstract classes of danger called \textit{criticality phenomena} and their underlying causal relations for a class of systems.

The Phenomenon-Signal Model (PSM) considers information as the basis of behavior in order to examine the flow of information between traffic participants \cite{beck2022phenomenon} and is, therefore, inherently tied to the investigation of causal questions.
While the criticality analysis seeks a causal understanding of the emergence of criticality, the PSM's main goal is to derive a valid target behavior for ADSs.

For the validation of sensor models, Linnhoff et al. \cite{linnhoff2021} introduce the \textit{Perception Sensor Collaborative Effect and Cause Tree} (PerCollECT) where they model, as an informal graph, the effect of system-independent causes on the active perception of sensor models.
While providing crucial information about the causal assumptions on the sensor model, the relevance estimation is expert-based and therefore inherently qualitative.
In contrast to our approach, PerCollECT does not provide a formal basis for the definition of nodes and edges grounded in probability theory.
However, in \cite{linnhoff2022}, Linnhoff et al.\ measure the influence of environmental conditions on automotive lidar sensors.
The data measured in this approach could be advantageously combined with a causal structure to derive the CPDs necessary to obtain a causal model.

In recent work, Maier et al. \cite{maier_bayesiansafety_2022} sketch their ideas how causal models could be used for automotive safety, in general, and scenario-based testing, in particular. Their approach shows similarities to the process for the criticality analysis. 
In contrast to these abstract sketches, the work at hand takes the use of causal models for automotive safety to a concrete level of application.

\section{Causal Relations as Semi-Markovian Causal Models}
\label{sec:causal_relations}

On the basis of Pearl's causal inference \cite{pearl2009causal}, we propose the following approach to achieve a causal understanding of a CP and its influence on criticality corresponding to the four aspects \textbf{Q1}-\textbf{Q4} outlined in \autoref{sec:preliminaries}. 
As a main idea we build on Pearl's causal theory in order to provide a formal definition of a causal relation enabling a comprehensive analysis of causality for automated driving.

\subsection{Criticality Phenomena and Criticality Metrics in Terms of Causal Theory}
\label{subsec:cp_phi_causal_theory}

The language of causal theory requires an adequate definition of the different terms. 
Firstly, we need to refine the notion of CP which is defined within the criticality analysis as \textit{a concrete influencing factor in a scenario associated with increased criticality} \cite[Definition~2]{neucrit21}.
Expressing a CP in terms of a causal theory means that it has to be described using random variables.  
In particular, when analyzing the causal influence of a CP on measured criticality, the do-calculus prescribes the investigation of an atomic intervention $\Do{X=x}$ for a suitable random variable $X$ and a value $x$.
Therefore, the CP has to occur as a single value of a random variable -- not as the random variable itself. 
Accordingly, we introduce a binary random variable $X$ with value range $\Ima(X) = \lbrace \CP, \nCP\rbrace$ where the value $\CP$ models the phenomenon and $\nCP$ its negation.

As a simple example, consider the phenomenon \textsc{heavy rain} ($=\CP$), roughly understood a strong form of liquid precipitation ($=X$). 
One possible formalization of the random variable $X$ could be based on the precipitation rate. 
By determining a threshold $\alpha$ for the precipitation rate, $X$ becomes a binary random variables that assumes the value $\CP$ = $\textsc{heavy rain}$, if the precipitation rate is higher than $\alpha$ and $\nCP$, if it at most $\alpha$. Generally, a formalization of both the variable (in the example, the precipitation rate) and the value $\CP$ representing the CP (in the example, $> \alpha$) shall be carefully based on design decisions regarding its relevancy in the criticality analysis as well as the practicability of its measurability in terms of basic concepts (such as precipitation) \cite{westhofen2022ontologies}. 
In some cases, domain standards and regulations are available to guide these decisions, but even in such cases the semantics might vary a lot. 
For \textsc{heavy rain} there exist several thresholds and different units characterizing it: 
The American Meteorological Association defines \textsc{heavy rain} as a precipitation in the form of liquid water drops with a maximum rate over $7.6$mm in sixty minutes or more than $0.76$mm in six minutes\footnote{\url{https://glossary.ametsoc.org/wiki/Rain}}.
In contrast, Germany's national meteorological service DWD defines \textsc{heavy rain} as a precipitation of more than $10 $mm in sixty minutes or more than $ 1.7$mm in ten minutes\footnote{\url{https://www.dwd.de/DE/service/lexikon/Functions/glossar.html?lv2=101812&lv3=101906}}.
Note that there are also other formalizations possible that are not using the basic concept of precipitation, e.g. based on the expected probability of occurrence.

As to investigate causal effects it is crucial that the value range of the random variable $X$ associated with a CP is feasibly measurable in the real world and preferably also in simulation. 
Hence, the value range and its unit need to be chosen carefully. 
For a unified semantics, the formalization chosen for a CP should be denoted in the respective domain ontology.
If appropriate data or measurement standards are already known at that point of the investigation, the value range and its unit should be specified accordingly.
The actual formalization of the CP can be shifted to a later point in time when the causal structure will be combined with data, however, its general measurability is mandatory. 
Note that the chosen formalization influences the causal analysis -- in the worst case, e.g. choosing an extremely low precipitation rate of \textsc{heavy rain}, causal effects might vanish. 
It can hence be necessary to refine the formalization of the CP \cite[Section V. A. 5)]{neucrit21}.

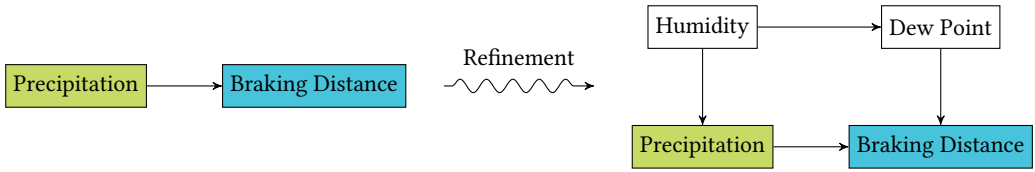
\begin{figure}
	\centering
	\usetikzlibrary{arrows, decorations.pathmorphing, calc}
\small
\begin{tikzpicture}[
		->,
		> = stealth',
		decoration = {snake, pre length=3pt,post length=7pt},
		every node/.style={minimum height=0.56cm},
	]
	\node[draw, align=center, fill=SpringGreen] (1) {Precipitation};
	\node[draw, align=center, right=1cm of 1, fill=SkyBlue] (2) {Braking Distance};
	\draw[->, > = stealth'] (1) edge (2);
	
	\node[draw, align=center, right=3cm of 2, yshift=-0.8cm, fill=SpringGreen] (3) {Precipitation};
	\node[draw, align=center, right=1cm of 3, fill=SkyBlue] (4) {Braking Distance};
	\node[draw, align=center, above=1cm of 3] (5) {Humidity};
	\node[draw, align=center, above=1cm of 4] (6) {Dew Point};
	\draw[->, > = stealth'] (3) edge (4);
	\draw[->, > = stealth'] (5) edge (3);
	\draw[->, > = stealth'] (5) edge (6);
	\draw[->, > = stealth'] (6) edge (4);
	
	\path[decorate, draw=black] ($(2.east)+(0.5cm, 0.0cm)$) -- node[above=0.1cm] {Refinement} ($(3.west)-(0.5cm, -0.8cm)$);
\end{tikzpicture}
	\Description{Exemplification of the modeling process for a causal structure with criticality phenomenon and metric.}
	\caption{Exemplification of the modeling process for a causal structure with criticality phenomenon and metric.}
	\label{fig:example_causal_relation_modeling}
\end{figure}

In order to measure criticality we employ so-called \textit{criticality metrics}, i.e.\ functions that estimate certain aspects of criticality in a scene or scenario \cite{westhofen2022metrics}. 
For the remainder of this work we denote  a criticality metric by $\varphi$.
As Pearl's causal theory argues about certain expected values, we further assume that $\varphi$'s codomain is subsumed in $[0,\infty)$, with higher outcomes implying increased criticality.
Thus, we assume criticality metrics to be of the form $\varphi : \mathcal{S} \rightarrow [0, \infty)$, quantifying the corresponding aspects of criticality for the investigated traffic situation $S \in \mathcal{S}$.
As each phenomenon might affect different aspects of criticality, the selection of the metric heavily impacts the causal analysis. 
Consequently, we need to evaluate possible metrics for every CP individually. 
It can be advantageous and sometimes even necessary to investigate the causal effects of a CP on multiple metrics in order to get a broader picture of its influence on criticality.
In \cite[Section~6]{westhofen2022metrics}, Westhofen et al.\ provide guidance on selecting suitable criticality metrics for the task at hand based on a review of the state of the art.
In the example of \textsc{heavy rain}, the braking distance of the ADS-operated vehicle might be a simple candidate metric to measure its effect on criticality.
The modeling process of a causal structure for $\CP=$ \textsc{heavy rain} $\in \Ima(X)$ with $X$ = precipitation and $\varphi=$ braking distance is illustrated by \autoref{fig:example_causal_relation_modeling}.
There, in a first refinement step, two variables hypothesized to be relevant for the causal effect of $X$ on $\varphi$ are added to the graph as nodes with the respective edges.

\subsection{Requirements on and Principles for Modeling Causal Relations}
\label{subsec:modeling_principles}
In the previous subsection, we framed criticality phenomena and metrics in causal language and gave an idea on the modeling and refinement of causal structures, as illustrated by \autoref{fig:example_causal_relation_modeling}. 
In this subsection, we look at what else is needed for a formal definition of causal relations besides a causal structure containing a CP and a metric and detail out the modeling process.

The overarching goal is to derive a causal structure that can be combined with data such that the causal effect of the investigated phenomenon on the measured criticality is identifiable in terms of the do-calculus. 
Essentially, the causal relations in the criticality analysis correspond to the concept of causal structures defined by Pearl, cf.~\autoref{subsec:intro_causal_models}.
However, there are additional requirements for the application of causal theory in the criticality analysis.
As each CP defines a broad class of scenarios -- the set of all scenarios where the CP is present -- the process of modeling everything that could be relevant to the CP's causal relation explicitly leads to arbitrarily complex causal structures.
Due to this issue, it appears natural to partition the class of scenarios for a given CP into subclasses which can be modeled more efficiently.
Pearl describes as \textit{context} a collection of assumptions and constraints that are assumed to be true and that every probability distribution is implicitly conditioned on \cite[pp.4-5]{pearl_2009}.
Based on this idea, we introduce a definition of a context for causal structures that represents a subset of the class of scenarios by defining constraints on and the existence of individuals in the scenario class for further reference in the graph structure.

\begin{definition}
	\label{def:context}
	The \textbf{context of a causal structure} is a set of statements about the existence of and constraints on the individuals contained in a suitable ontology.
\end{definition}

For the automotive domain, there exist multiple promising ontologies that might be used to define contexts of causal structures, e.g.\ A.U.T.O. \cite{westhofen2022ontologies} or ASAM OpenXOntology \cite{asamOpenXOntology}.
The 6-layer model (6LM) as presented in \cite{scholtes2021layer} may be used to structure the set of statements provided by the context.
Note that, generally, an increased number of assumptions and constraints in the context lead to simpler causal structures at the cost of reducing its universality and thus its coverage.
The context serves as existential basis for building the causal structure for the CP, which includes modeling its emergence as well as its influence on a criticality metric.
Therefore, combining the causal structures from \autoref{subsec:cp_phi_causal_theory} with \autoref{def:context}, we arrive at the following:

\begin{definition}
	\label{def:causal_relation_new}
	A \textbf{causal relation} for a CP $\CP$ is a tuple $(\mathfrak{S}, \mathfrak{C})$ with respect to a suitable traffic domain ontology $\mathcal{O}$ where $\mathfrak{S}=(V \cup U,E)$ is a causal structure and $\mathfrak{C}$ is a context such that there exists
	\begin{itemize}
		\item[(i)] a binary random variable $X$ with $\Ima(X) = \{ \CP, \nCP \}$ corresponding to a node in $V$ and
		\item[(ii)] a criticality metric $\varphi : \mathcal{S} \rightarrow [0,\infty)$ corresponding to a node without outgoing edges in $V$,
		\item[(iii)] where the variables in $U$ are the exogenous variables corresponding to error terms,
		\item[(iv)] the variables in $V$ are defined on properties of the individuals in $\mathcal{O}$, and
		\item[(v)] the context $\mathfrak{C}$ is defined as in \autoref{def:context} with respect to $\mathcal{O}$.
	\end{itemize}
\end{definition}
\begin{remark}[Causal Relation]\
	\label{rem:causal_relation_new}
	\begin{itemize}
		\item regarding (ii): in principal, it is possible to model more than one criticality metric in the causal structure, but this case can easily be reduced to the single metric case by considering several causal relations - each with only one metric in the graph 
		\item regarding (ii): the chosen criticality metric $\varphi$ has to be suitable for measuring the aspects of criticality associated with $\CP$
		\item regarding (ii): the causal structure's dependencies cause the metric to become a random variable
		\item regarding (iii): formally, there exists an exogenous variable for every endogenous variable, however, these errors terms are usually not modeled explicitly
	\end{itemize}
\end{remark}

\begin{figure}[htb!]
	\centering
	\scalebox{0.7}{\begin{tikzpicture}[]
	\node[align=left, draw=black] 				        (a) {$V_1$};
	\node[align=left, right=1cm of a,draw=black]	    (b)	{$V_2$};
	\node[align=left, right=1cm of b,draw=black]     	(c)	{$V_3$};
	\node[align=left, right=1cm of c,draw=black]	    (d)	{$V_4$};
	\node[align=left, below right=0.5cm and 0.25cm of a, draw=black] (e) {$V_5$};
	\node[align=left, right=1cm of e, draw=black, fill=SpringGreen] (f) {$X$};
	\node[align=left, right=1cm of f, draw=black] (g) {$V_6$};
	\node[align=left, below right=0.5cm and 0.175cm of e, draw=black] (h) {$V_7$};
	\node[align=left, right=1cm of h, draw=black] (i) {$V_8$};
	\node[align=left, below right=0.5cm and 0.25cm of h, draw=black,  fill=SkyBlue] (k)	{$\varphi$};
	
	\node[align=left, above=0.5cm of d,draw=black] 	(l) {$V_4 = g_4(\mathtt{a.p_1}, \dots, \mathtt{a.p_n}, \mathtt{b.p_2}, \dots, \mathtt{b.p_m}, \mathtt{c.p_1}, \dots, \mathtt{c.p_k})$};
	
	\draw (l.south east) edge[dotted] node{} (d.north east);
	\draw (l.south west) edge[dotted] node{} (d.north west);
	\draw[->]
	(a) edge[] node{}	(e)
	(b) edge[] node{}	(e)
	(b) edge[] node{}	(f)
	(c) edge[] node{}	(f)
	(c) edge[] node{}	(g)
	(d) edge[] node{}	(g)
	(d) edge[] node{}	(f)
	(f) edge[] node{}	(h)
	(f) edge[] node{}	(i)
	(g) edge[] node{}	(i)
	(e) edge[] node{}	(i)
	(e) edge[] node{}	(h)
	(h) edge[] node{}	(k)
	(i) edge[] node{}	(k);

	\node[cloud, draw, fill = gray!10, align=left, inner sep=0pt, aspect=4, cloud puffs=15, cloud puff arc=100] (context) at (11,-1.3) {Context $\mathfrak{C}$:\\$\bullet$ \texttt{a} exists and is of type $\mathtt{C_1}$ \\$\bullet$ \texttt{b} exists and is of type $\mathtt{C_1}$, but does not have property $\mathtt{p_1}$\\ $\bullet$ \texttt{c} exists and is of type $\mathtt{C_2}$\\$\bullet$ It has to hold that $\mathtt{a}.\mathtt{p_2} < \mathtt{b}.\mathtt{p_2}$.};
\end{tikzpicture}}
	\Description{An exemplary causal relation with causal structure $\mathfrak{S}$ on the left, error terms omitted, and context $\mathfrak{C}$ on the right. $X$ takes the values $\Ima(X)=\{\CP,\nCP\}$ and $\varphi$ is a criticality metric. $\mathtt{C_1}$ and $\mathtt{C_2}$ are classes and \texttt{a}, \texttt{b} and \texttt{c} are individuals in the domain ontology. The inputs of the random variable $V_4$ are indicated by a function $g_4$ taking as inputs the properties of the individuals in the ontology.}
	\caption{An exemplary causal relation with causal structure $\mathfrak{S}$ on the left, error terms omitted, and context $\mathfrak{C}$ on the right. $X$ takes the values $\Ima(X)=\{\CP,\nCP\}$ and $\varphi$ is a criticality metric. $\mathtt{C_1}$ and $\mathtt{C_2}$ are classes and \texttt{a}, \texttt{b} and \texttt{c} are individuals in the domain ontology. The inputs of the random variable $V_4$ are indicated by a function $g_4$ taking as inputs the properties of the individuals in the ontology.}
	\label{fig:simple_cr}
\end{figure}
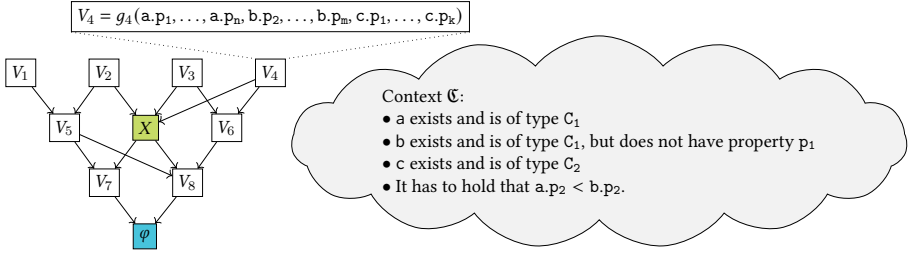

As to illustrate \autoref{def:causal_relation_new} and \autoref{rem:causal_relation_new}, consider the example of \autoref{fig:simple_cr}, which depicts a simple, generic causal relation.
The causal structure contains $10$ nodes including the random variable $X$ near the center of the graph and a criticality metric $\varphi$ at the cusp.
In this example, it is assumed that the variables $V_2, V_3$ and $V_4$ causally influence $X$, while $X$ influences $\varphi$ through $V_7$ and $V_8$.
These random variables encode properties of the individuals $a, b$ and $c$ as defined in the context $\mathfrak{C}$.
Taken together, the causal structure and the context of \autoref{fig:simple_cr} define a causal relation.

\paragraph{Modeling of Causal Relations}
As to build the graph structure, the relevant 'propositions on traffic scenarios' related to the CP and criticality metric need to be represented as variables structured in a directed graph.
In this graph, each edge represents the possibility of a direct functional relationship between the connected variables, i.e.\ a causal assumption, without necessarily being explicit.

As the causal structure serves as a blueprint for a causal model the variables should be expandable to random variables.
Hence every variable has to be investigated regarding the possibility to assign a value range including a unit.
This is necessary to assure that it is in general possible to combine the variables with data even if the variables are not measurable in a concrete experimental or non-experimental setup. 
Typically the inability of assigning a value range is due to an inadequate definition of the variable or an inappropriate abstraction level. 
In that case we have to decide whether it is sufficient to keep the variable as an unobserved background factor, a so-called error term, or if further concretization of the variable and the causal structure is required.

Note that the application of the do-calculus is defined on discrete variables. 
In fact, many scenario variables, including the criticality metrics, are typically connected to a continuous distribution. 
However, even if the variables are not discrete in reality, some discretization level must be found during the modeling process that sufficiently captures the features, yet does not require unfeasible amounts of data to adequately estimate the distribution.
Nevertheless, it must always be considered that discretization implies error terms further down the line and that these error terms may only be hold negligible or even be estimated for rather well-behaved variables \cite{neurfund20}.

\paragraph{Temporal Aspects}
We remark that every causal assumption implicates a certain temporal dependency between the corresponding nodes, as the cause is assumed to precede the effect. 
However, modeling the causal structure without explicit consideration of the temporal aspects is a form of abstraction which can lead to cycles in the structure, as the dependencies of scenario properties might vary at different points of time \cite[Chapter 10]{peters2017}.
Therefore, the question if we need to model temporal dependencies explicitly is mainly based on the question if there exists an adequate abstraction level.
In this work, we focus on modeling causal dependencies without temporal aspects and leave
the investigation on time-dependent structures for future research. The whole causal structure should be refined until there are at least no cycles in the structure in order to obtain a semi-Markovian model.
Further, every underlying causal assumption should be examined carefully.

\paragraph{Plausibilization of Causal Relations} 

The quantitative assessment of causal relations requires the combination of the causal structure with data. 
As we can not expect that a dataset covers every variable, we introduce the following definition allowing the usage of partial data:

\begin{definition}	
	Let $\CR = ((V,E), \mathfrak{C})$ be a causal relation and $N\subseteq V$ a subset of nodes that contains a node corresponding to a variable $X$ with $\Ima(X) = \{\CP, \nCP\}$.
	Given a dataset $D$ we call $\CR$ \textbf{partially instantiated w.r.t. $N$ by $D$} if the CPDs of  each variable contained in $N$ are instantiated by $D$.
	Further, if $N$ additionally contains a node $\varphi$ representing the criticality metric and at least one adjustment set for the causal effect of $X$ on $\varphi$, we call a causal relation $\CR$ \textbf{instantiated w.r.t. $N$ by $D$}.
\end{definition}

In order to assure that the causal understanding of a CP specified by a causal relation is valid, the criticality analysis includes an evaluation whether the causal relation models the emergence of the CP and its influence on criticality sufficiently (cf.~steps 5. and 6. in \autoref{fig:process}).
For this, we propose an expert-based approach based on statistical hypothesis testing.
Here, we assess the modeling quality of a causal relation $\CR$ w.r.t. a dataset by employing some functions $\kappa$ with values in $[0,\infty)$ for which values closer to zero imply that the model fits to the data.
These judging functions can be used as test statistic to decide based on a dataset $D$ whether the hypothesis -- $\CR$ does not explain the emergence of the CP or its influence on the criticality metric sufficiently -- should be rejected or not.
If for a selected family of datasets the hypothesis can be rejected with a significance level $\alpha>0$ or if the experts decide that those datasets that do not support a rejection are insignificant, we call $\CR$ plausibilized w.r.t. $\kappa$ with significance level $\alpha$.
To achieve a comprehensive plausibilization process, investigations regarding different judging functions might be necessary.
\autoref{subsec:modeling_quantities} introduces some functions that may be used for this plausibilization.
However, the construction of corresponding test statistics, including the estimation of critical regions, is left for future research.

\subsection{Quantification of Modeling Quality}
\label{subsec:modeling_quantities}

Since we now clarified a common understanding of the type of variables that represent the CP and the criticality metric, we can now define the main quantities suggested for the iteration cycle.
The indicators proposed in this section shall provide an insight into the causal queries \textbf{Q1}, \textbf{Q2} and \textbf{Q3} and thus will be called causality indicator functions.
While \textbf{Q4} is not directly addressed in this section, it can be approached using the quantities defined in this section as well.

In order to address \textbf{Q2}, the following quantities measure the increase in criticality caused by a phenomenon using a difference of expected values (respectively a ratio):
\begin{definition}
	Let $\mathfrak{S}=((V,E),\mathfrak{C})$ be a causal relation and $D$ a dataset s.t. $\mathfrak{S}$ is instantiated w.r.t. $N$ by $D$ for a suitable set of nodes $N \subseteq V$.
	The \textbf{average (resp.\ relative) causal effect of} $X$ \textbf{on} $\varphi$ are
	\begin{align*}
		\mathit{ACE}(X, \varphi) &\coloneqq E(\varphi\mid \Do{X=\CP}) - E(\varphi\mid\Do{X=\nCP})\,, \\
		\mathit{RCE}(X, \varphi) &\coloneqq \frac{E(\varphi\mid \Do{X=\CP})}{E(\varphi\mid\Do{X=\nCP})}\,.
	\end{align*}
	\label{def:abs_and_rel_causal_effect_phenomenon_to_criticality}
\end{definition}
The causality indicator functions $\mathit{ACE}$ and $\mathit{RCE}$ are quite intuitive and generalize the average (resp.\ relative) treatment effect to non-experimental data using the do-calculus.

The causal query \textbf{Q3} evaluates whether the chosen criticality metric recognizes the effect of the CP and if so, whether the CP has a significant causal impact on the criticality. Therefore, we propose a quantity which incorporates the proportion of the measured criticality outside of the CP:

\begin{figure}
	\footnotesize
	\centering
	\begin{minipage}{0.37\linewidth}
		\centering
		\begin{subfigure}{\linewidth}
			\centering
			\includegraphics[width=0.75\linewidth]{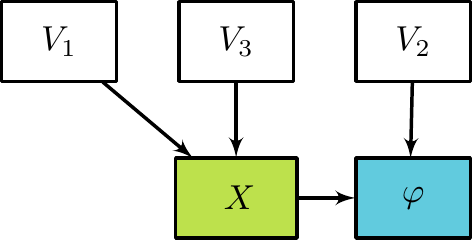}
			\subcaption{Assumed Reality}
			\label{fig:rho_models_a}
		\end{subfigure}
		\vspace{0.35cm}\dotfill\vfill
		\begin{subfigure}{\linewidth}
			\centering
			\includegraphics[width=0.75\linewidth]{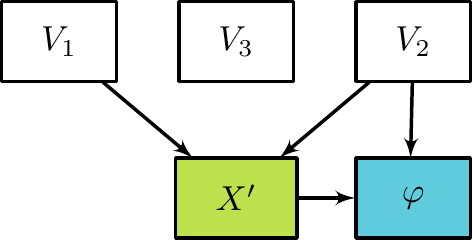}
			\subcaption{Model of assumed reality}
			\label{fig:rho_models_b}
		\end{subfigure}
	\end{minipage}
	\hfill
	\begin{minipage}{0.6\linewidth}
		\begin{tabularx}{0.32\linewidth}{Xr}
			\toprule
			$V_1=$~Summer & 0.5 \\
			$V_1=$~Winter & 0.5 \\
			\bottomrule
		\end{tabularx}
		\hfill
		\begin{tabularx}{0.32\linewidth}{Xr}
			\toprule
			$V_2=$~Slow & 0.6 \\
			$V_2=$~Fast & 0.4 \\
			\bottomrule
		\end{tabularx}
		\hfill
		\begin{tabularx}{0.32\linewidth}{Xr}
			\toprule 
			$V_3=$~Oceanic & 0.6 \\
			$V_3=$~Cont. & 0.4 \\
			\bottomrule
		\end{tabularx}
		\hfill
		\begin{tabularx}{\linewidth}{Xcccc}
			\toprule
			$V_1$ & \multicolumn{2}{c}{$V_1=$\,Summer} & \multicolumn{2}{c}{$V_1=$\,Winter} \\
			$V_3$ & $V_3=$~Oceanic & $V_3=$~Cont. & $V_3=$~Oceanic & $V_3=$~Cont. \\ 
			$X=\nCP$ & 0.4 & 0.2 & 0.3 & 0.4 \\
			$X=\CP$ & 0.6 & 0.8 & 0.7 & 0.6 \\
			\bottomrule
		\end{tabularx}
		\begin{tabularx}{\linewidth}{Xcccc}
			\toprule
			$V_1$ & \multicolumn{2}{c}{$V_1=$~Summer} & \multicolumn{2}{c}{$V_1=$~Winter} \\
			$V_2$ & $V_2=$~Slow & $V_2=$~Fast & $V_2=$~Slow & $V_2=$~Fast \\ 
			$X'=\nCP$ & 0.4 & 0.2 & 0.3 & 0.4 \\
			$X'=\CP$ & 0.6 & 0.8 & 0.7 & 0.6 \\
			\bottomrule
		\end{tabularx}
		\begin{tabularx}{\linewidth}{Xcccc}
			\toprule
			$X, X'$ & \multicolumn{2}{c}{$X, X'=\nCP$} & \multicolumn{2}{c}{$X, X'=\CP$} \\
			$V_2$ & $V_2=$~Slow & $V_2=$~Fast & $V_2=$~Slow & $V_2=$~Fast \\ 
			$\varphi=$~Short & 0.6 & 0.1 & 0.8 & 0.3 \\
			$\varphi=$~Long & 0.4 & 0.9 & 0.2 & 0.7 \\
			\bottomrule
		\end{tabularx}
	\end{minipage}
	\Description{Exemplary models for demonstrating the suggested causality indicator functions. We have $\varphi$ = braking distance, $X, X'$ = precipitation, $V_1$ = season, $V_2$ = ego forward velocity and $V_3$ = climate region.}
	\caption{Exemplary models for demonstrating the suggested causality indicator functions. We have $\varphi$ = braking distance, $X, X'$ = precipitation, $V_1$ = season, $V_2$ = ego forward velocity and $V_3$ = climate region.}
	\label{fig:rho_models}
\end{figure}

\begin{definition}
	Let $\mathfrak{S}=((V,E),\mathfrak{C})$ be a causal relation and $D$ a dataset such that $\mathfrak{S}$ is instantiated w.r.t. $N$ by $D$ for a suitable set of nodes $N \subseteq V$.
	We further require the criticality metric $\varphi$ in $N$ to fulfill $E(\varphi | \Do{X = \nCP}) \leq E(\varphi | \Do{X = \CP})$.
	The \textbf{extent of the explanation of} $\varphi$ \textbf{by} $X$ is then
	\begin{equation*}
		\sigma(X, \varphi) \coloneqq 1 - \frac{E(\varphi | \Do{X=\nCP})}{E(\varphi)}\,, \quad \text{ denoted } \CP \models_\sigma \varphi\,.
	\end{equation*}
	\label{def:explanation_criticality_by_phenomenon}
\end{definition}

This causality indicator function formalizes the causal query \textbf{Q3} by evaluating the strength of the paths from $X$ to $\varphi$ within the causal relation.
Note that the denominator is a purely stochastic and $\sigma$ therefore compares a causal with a purely observational quantity.

Moving on to \textbf{Q1}, we discuss and define several indicators that allow insight into the emergence of the CP. 
As a first simple guess, one might think that a pairwise comparison of distributions of a subset of nodes between models is a feasible approach. In the special case of comparing the distribution of $X$ in the model with the distribution of $X$ in reality, we denote this with $\rho_1 = \mathit{KL}(P^{X_M}|P^{X_R})$, where here and in the following $\mathit{KL}$ denotes the Kullback-Leibler divergence \cite{mackay2003}.
However, it is very likely that even though one got the distribution of occurrence of CP correct, the dependencies leading to the emergence of CP are incorrect.
To demonstrate this, we present an example of two causal models of the CP \textsc{heavy rain}, where one of the models is declared as reality (\autoref{fig:rho_models_a}) while the other is a model that an expert came up with (\autoref{fig:rho_models_b}), trying to model said reality.
According to \autoref{subsec:cp_phi_causal_theory} we model the CP as a binary variable $X$, where $CP$ is equal to \textsc{heavy rain} and define $\varphi$ as the braking distance with the possible values $\{\emph{Short}, \emph{Long}\}\,\hat{=}\,\{0, 1\}$.
Since the parents of $X$ in the two models are not equal, the index on $X$ solely clarifies which CPDs are used.
Further, with the other variables we denote $V_1$ as the season, $V_2$ as the ego forward velocity and $V_3$ is the climate region that is equally distributed to $V_2$ for this example.
Computing the aforementioned quantities, we obtain $ACE = 0.2$, $RCE = 1.5$, $\sigma = 0.1416$ and $\rho_1 = 0$.
Clearly, $\rho_1=0$ does not identify the insufficient modeling quality here. As a first improvement regarding \textbf{Q1} we suggest an indicator that incorporates stochastic dependencies among the variables:

\begin{definition}
	Let $\mathfrak{S}=((V,E),\mathfrak{C})$ be a causal relation, $N \subseteq V$ a set of nodes, $D_1$ and $D_2$ datasets containing real world data and data from the model, respectively, such that $\mathfrak{S}$ is partially instantiated w.r.t. $N$ by $D_1$ and by $D_2$.
	Further, we denote with $P$ and $Q$ joint probability distributions for the variables in $N$, obtained by estimating the distribution from $D_1$ and $D_2$, respectively.
	We then define the \textbf{extent of explanation of the observational emergence of $\CP$ w.r.t. $N$} as the Kullback-Leibler divergence of the distributions $P$ and $Q$, i.e. $\rho_2^N = KL(P(N)~|~Q(N))$.
\end{definition}

As this causality indicator rewards us with a combined causality indicator function for the emergence of the CP, that also includes the strenghts of dependencies in both models, we obtain $\rho_2 = 0.0141 > 0$ for the example of \autoref{fig:rho_models}.
However, due to the associational character of the metric, it may only infer what Peters et al. termed \textit{observational equivalence} \cite[Def.~6.47]{peters2017} and is thus not able to implicate equivalence of interventions.
To tackle this issue, we introduce a quantity using the concept of \emph{causal influence} as defined by Janzing et al.\ \cite[Def.~2]{janzing2013}.

\begin{definition}
	Let $\mathfrak{S}=((V,E),\mathfrak{C})$ be a causal relation, $N \subseteq V$ a set of nodes, $D_1$ and $D_2$ datasets containing real world data and data from the model, respectively, such that $\mathfrak{S}$ is partially instantiated w.r.t. $N$ by $D_1$ and by $D_2$.
	We denote for $n \in V$ by $\out(n) \subseteq E$ the set of outgoing edges of the node $n$ and denote the causal influence \cite[Def. 2]{janzing2013} of $n$ computed using $D_1$ by $\mathfrak{I}^R_{\out(n)}$ and computed from $D_2$ by $\mathfrak{I}^M_{\out(n)}$.
	Finally, we obtain the vector representation of $\rho_3^N$ as
	\begin{equation*}
	\rho^N_{3,n} = \mathfrak{I}^R_{\out(n)} - \mathfrak{I}^M_{\out(n)}~\text{for all}~n \in N \setminus X
	\end{equation*}
	and take the norm to define the \textbf{extent of explanation of the emergence of $\CP$ w.r.t. $N$ using causal influence} as the scalar value $\rho_3^N = \|(\rho^N_{3,n})_{n \in N \setminus X}\|_2$.
\end{definition}

Computing this quantity for the example of \autoref{fig:rho_models} and with $N = \{V_1, V_2, X\}$, we obtain $\rho_3^N = 0.0181 > 0$.
Another consideration regarding these causality indicators is that it might not only be of interest whether the dependencies of $X$, but also the dependencies of ancestors of $X$, are modeled correctly. 
In those cases we would consider the extent of explanation of the emergence of $\CP$ to be an aggregate of multiple values computed in analogy to the definitions of the $\rho_i$.

Note that we define $\rho_1$, $\rho_2$ and $\rho_3$ as causality indicator functions giving certain directions to experts about the quality of the model and that they should not be construed as decision metrics. 
As mentioned before, function $\rho_2$ itself is purely associational and thus can only identify dependencies, not causalities. 
However, since correct dependencies are a necessary condition for correctly modeled causalities, computing this indicator contributes to the iterative plausibilization.
Following Peters et al.\ \cite{peters2017}, observational equivalence is a prerequisite for interventional equivalence.
Furthermore, in practice it is beneficial to be able to compare only subsets of the vertices of a graph, especially during the design phase, which the presented causality indicator functions allow for.
Note that the expressive power of the presented causality indicator functions is based on data quality in the comparison dataset, since errors due to insufficient data are included.
This underlines the notion of the indicator functions not being decision metrics but rather guidance for the expert.

Finally, let us remark that there is much room for improving such distance measures between probability distributions.
Particularly promising are metrics that provide constructive feedback, e.g. in the form of witness functions, as well as distance metrics that do not require estimation of the probability distributions, e.g.\ maximum mean discrepancy \cite{gretton2012}.

\section{Causal Relation for Reduced Coefficient of Friction}
\label{sec:example}

\begin{wraptable}{r}{0.5\linewidth}
	\scriptsize
	\caption{Examplary context for the causal relation for the CP \textsc{Reduced Coefficient of Friction}.}
	\begin{tabularx}{\linewidth}{p{3.1cm}X}
		\toprule
		{\bf Layer} & {\bf Property} \\
		(L1) Road Network and Traffic Guidance Objects & A road network shall exist and shall consist of either a curved road or a junction. \\
		(L2) Roadside Structures & Roadside structures may exist and are not further constrained. \\
		(L3) Temporary Modifications of (L1) and (L2) & There shall be no temporary modifications to layer 1 and 2. \\
		(L4) Dynamic Objects & An ego vehicle shall exist. No other vehicle relevant to the ego vehicles actions and behavior shall exist. \\
		(L5) Environmental Conditions & Environmental conditions shall exist and remain unconstrained. \\
		(L6) Digital Information & Digital information might exist, but shall remain unconstrained. \\ \bottomrule
	\end{tabularx}
	\label{tab:context_coefficient_of_friction}
\end{wraptable}

In this section, we apply the methods of \autoref{subsec:cp_phi_causal_theory} and \autoref{subsec:modeling_principles} to the example CP \textsc{reduced coefficient of friction} and its causal relation.
Further explanations for the nodes regarding their interpretation and properties as a random variable can be found in \autoref{subsec:appendix_1}.
Following Reichenbachs common cause principle \cite{wronski2014}, it is possible for small time scales and detailed models of the reality (see \autoref{tab:example_models}) to model the causalities as a Markovian causal structure.
Since an underspecification of the context would lead to incomprehensible causal structures, we limit the causal relation to describe the objects mentioned in \autoref{tab:context_coefficient_of_friction}.
That means in particular, that even though other traffic participants may be present in the scene, they shall have no impact on more than one node in the causal structure for the model to be applicable.
While the inclusion of one or more traffic participants in the model is possible in general, it is likely to introduce cycles in the causal structure, requiring the usage of temporal modeling.

\begin{sidewaysfigure}
	\includegraphics[width=\textheight]{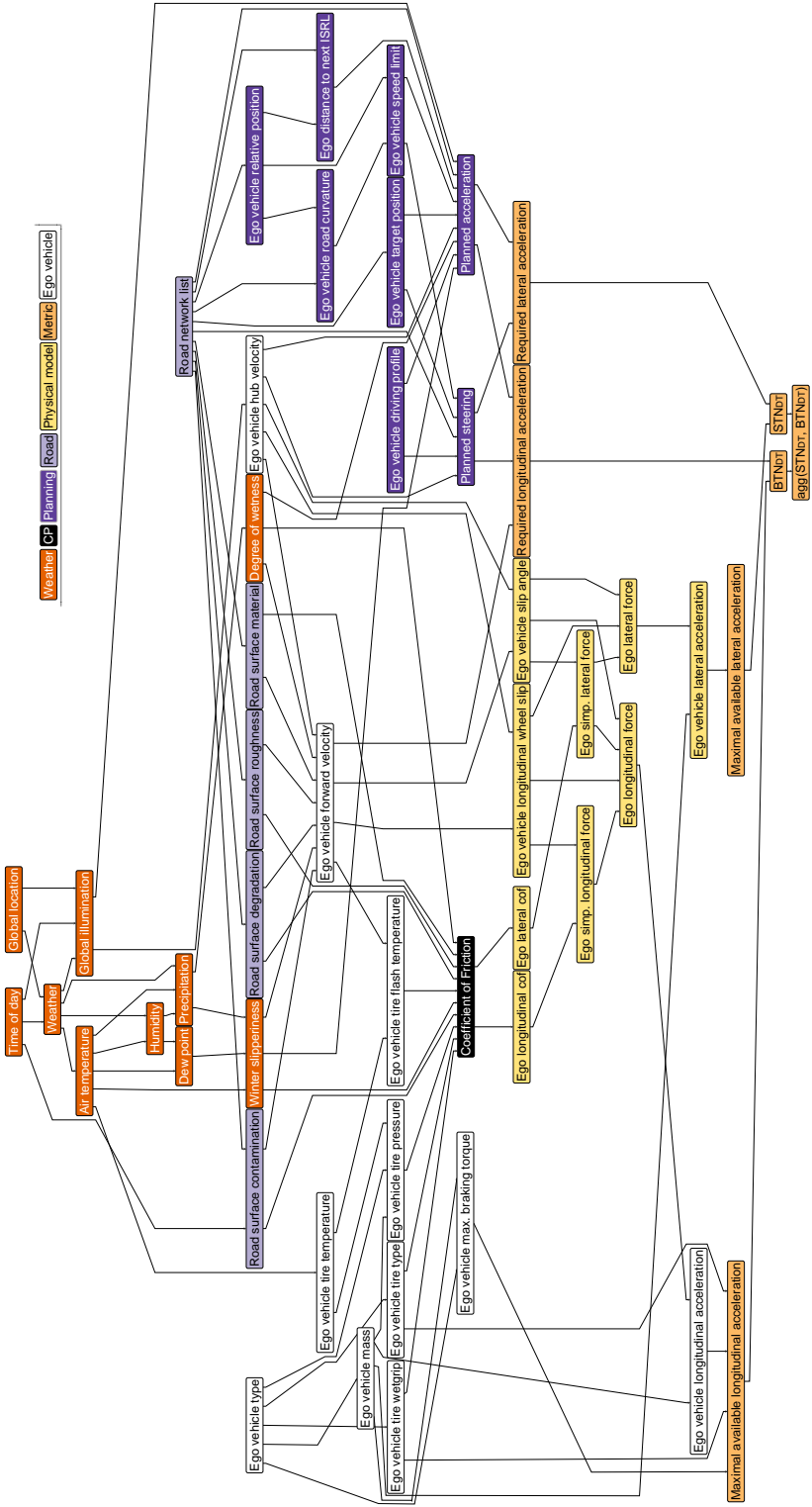}
	\Description{Causal structure for the CP \textsc{reduced coefficient of friction}.}
	\caption{Causal structure for the CP \textsc{reduced coefficient of friction}.}
	\label{fig:causal_relation_cof}
\end{sidewaysfigure}

For choosing a criticality metric $\varphi$, we conducted a suitability analysis as proposed in Westhofen et al. \cite[Section~6]{westhofen2022metrics}.
We derived the minimal requirements on the properties of the criticality metric as shown in \autoref{tab:requirement_on_metric} and obtained that the metrics TTB, TTR, TTS, $a_{lat, req}$, $a_{long, req}$, $a_{req}$, $\btn$, $\stn$, CPI, RSS-DS, TCI, P-MC, P-SMH, P-SRS and SP would be applicable.

\begin{wraptable}{l}{0.5\linewidth}
	\scriptsize
	\caption{Requirements on the criticality metric $\varphi$.}
	\begin{tabularx}{\linewidth}{lX}
		\toprule
		{\bf Property} & {\bf Requirement} \\
		Scenario type & Curved roads and intersections \\
		Validity & High \\
		Sensitivity & High \\
		Inputs & Available longitudinal and lateral dynamical control \\
		Specificity & Medium \\
		Reliability & High \\
		Output scale & Ordinal \\
		Subject type & Automation \\
		Run-time capability & None \\
		Target values & None \\
		Prediction model & No prediction necessary \\\bottomrule
	\end{tabularx}
	\label{tab:requirement_on_metric}
\end{wraptable}

However, since our use case does not specify the existence of another relevant traffic participant, we chose a modified combination of the BTN and STN as the metric for our example, that is computed with respect to a preassigned driving task.
Denoting with $\mathcal{T}$ the set of all possible $C^2$-trajectories of the ego vehicle, i.e. twice continuously differentiable functions $t \rightarrow {\bf p}(t) = (x(t), y(t))$ describing the time-dependent position of the ego vehicle, we then define the driving task (DT) as the subset of $\mathcal{T}$ in which every trajectory fulfills a given set of mobility, performance, safety and comfort goals. 
Further, we denote with $S$ a situation, $t_S$ the time of that situation and $t_H$ the time horizon, that mostly depends on how far into the future the driving task is predicted.
We then have
\begin{equation*}
	\alongreqdt(S) = \max\{a \leq 0~\mid~DT \cap \Gamma_{long,a} \neq \emptyset\}
\end{equation*}
with $\Gamma_{long,a} = \{\gamma \in \mathcal{T}\mid\;\forall\; t \in [t_S, t_S + t_H]\,:\,\ddot{\gamma}_{long}(t) \geq a \}$ denoting the subset of all trajectories that can be completed with a given longitudinal acceleration not smaller than $a$, and
\begin{equation*}
	\alatreqdt(S) = \min\{a \geq 0~\mid~DT \cap \Gamma_{lat,a} \neq \emptyset\}
\end{equation*}
with $\Gamma_{lat,a} = \{\gamma \in \mathcal{T}\mid \;\forall\;t \in [t_S, t_S + t_H]\,:\, |\ddot{\gamma}_{lat}(t)| \leq a\}$ being the subset of all trajectories that can be completed with a given absolute lateral acceleration of at most $a$.
By defining a function $\eta: \mathbb{R}^2 \rightarrow \mathbb{R}^2$ that assigns each point of the road network an available acceleration, we can write
\begin{equation*}
	\alongmin = \max\{\eta(\gamma(t))_{long}\mid\gamma \in DT, t \in [t_S, t_S + t_H]\}
\end{equation*}
and, respectively, for the available lateral acceleration
\begin{equation*}
	\alatmin = \min\{|\eta(\gamma(t))_{lat}|\mid\gamma \in DT, t \in [t_S, t_S + t_H]\}\,.
\end{equation*}
With these quantities we obtain the driving task dependent BTN resp. STN as
\begin{equation}
	\btndt(S) = \frac{\alongreqdt(S)}{\alongmin(S)}, \quad \stndt(S) = \frac{\alatreqdt(S)}{\alatmin(S)}\,.
\end{equation}

For the main example of \autoref{fig:causal_relation_cof} we suggest an aggregate of both $\btndt$ and $\stndt$ as criticality metric $\varphi$. The exact details of this aggregation, however, remain undefined as they would require calibration experiments which are out-of-scope for this work.
Since $\varphi$ is supposed to provide a notion of criticality in our study of the causal connections for a given CP, the criticality metric itself and the metrics direct components, e.g. the required longitudinal acceleration for $\agg(\btndt, \stndt)$ in \autoref{fig:causal_relation_cof}, shall not be part of possible adjustment sets.

\begin{wraptable}{r}{0.5\linewidth}
	\scriptsize
	\caption{Employed models during modeling of the CR for the CP \textsc{reduced coefficient of friction.}}
	\begin{tabularx}{\linewidth}{Xc}
		\toprule
		{\bf Scope of the model} & {\bf Source} \\
		Impact of tire (flash) temperature on coefficient of friction & \cite{Persson2006} \\
		Lateral/longitudinal aspects & \cite{Jansson2005} \\
		Validity of the friction ellipse & \cite{brach2011} \\ \bottomrule	
	\end{tabularx}
	\label{tab:example_models}
\end{wraptable}

These nodes will thus be set to unobserved during the qualitative analysis of the causal relation, directly removing them from the list of possible nodes included in adjustment sets.

\begin{table}
	\footnotesize
	\caption{Measurability of an exemplary adjustment set for the effect of the criticality phenomenon \textsc{Reduced Coefficient of Friction} on a criticality metric $\agg(\stndt,\btndt)$ based on the causal structure in \autoref{fig:causal_relation_cof}.}
	\label{tab:adjustment_cof}
	\begin{tabularx}{\linewidth}{p{3.9cm}X}
		\toprule
		{\bf Adjustment variable} & {\bf Measurability} \\
		Ego vehicle tire temperature & In-vehicle measurement with a sensor. \\
		Planned steering & Can be obtained from the planner component. \\
		Ego vehicle longitudinal wheel slip & Is provided by the electronic stability control (ESC).\\
		Ego vehicle tire wetgrip & Can be inferred from the tire imprint. \\
		Ego vehicle tire type & Can be inferred from the tire imprint. \\
		Planned acceleration & Can be obtained from the planner component. \\
		Ego vehicle tire pressure & In-vehicle measurement with a sensor. \\
		Ego vehicle forward velocity & May be obtained using global navigation satellite system (GNSS) sensors.\\
		Ego vehicle slip angle & Is provided by the electronic stability control (ESC).\\ \bottomrule
	\end{tabularx}
\end{table}

As to determine the possible adjustment sets for the causal effect of the CP \textsc{reduced coefficient of friction} on $\varphi$, the causal structure as presented in \autoref{fig:causal_relation_cof} was imported into the python library pgmpy \cite{ankan_pgmpy_2015} and the analysis resulted in the adjustment set listed in \autoref{tab:adjustment_cof}.
As a reminder, we can use these adjustment sets in the computation of causal effects as detailed in \autoref{eq:causal_effect_adjustmentset}.
The derivation of suitable adjustment sets thereby solely relies on the graph structure, which makes it a purely qualitative process, and helps to limit data collection requirements for subsequent quantitative examinations.

\section{Requirement Elicitation for Data Acquisition}
\label{sec:elicitation}

During the plausibilization of a causal relation for the derivation and assessment of SPs, the causality indicator functions answering the causal queries \textbf{Q1} and \textbf{Q2} have to be repeatedly estimated from data.
This section will give some first insight on the type of requirements that the inference of causal quantities based on the formulation of causal models pose onto the data acquisition processes both in reality and in simulation environments.
The data acquisition required for the iterative modeling of causal relations (cf.~steps 4.-7. in \autoref{fig:process}) will have to be done using real world data, since the validity and the demonstration of the validity of simulation environments for a given context is yet an open problem. 
Although major efforts have been put into researching the field of validation of simulation environments and simulation models \cite{riedmaier2020, rosenberger2019towards, rosenberger2020sequential}, there exists no well defined widely applicable validity metric (VM) with an attached threshold (VT), let alone models that fulfill such a requirement VM < VT, to provide a valid simulation.
Once the causal model is established and validly represented in the simulation, simulators can be employed to determine the effectiveness of the derived SPs (cf.~steps 8.-11. in \autoref{fig:process}).

\subsection{Real World Data}

Once an initial expert-based causal relation has been established, it is necessary for iterative plausibilization (steps 4.-6. in \autoref{fig:process}) of the model to collect data for the relevant variables, that are represented as vertices in the causal model, in order to evaluate the causality indicator functions.
In order to define the relevancy of the nodes regarding this data collection process we can make use of the underlying causal framework, that rewards us with an answer in terms of adjustment sets.
In our example in \autoref{sec:example} we have listed an adjustment set for the causal effect of the CP \textsc{reduced coefficient of friction} on an aggregated criticality metric $\agg(\stndt,\btndt)$. 
By collecting data for the variables that are part of the adjustment set, the CP and the criticality metric, we are able to compute both a combination of the $\rho_i$ as well as $\sigma$ for the refinement cycle.
However, this implies the requirement that every variable of the selected adjustment set can be recorded in reality.
Additionally, due to rare combinations of characteristics it might prove beneficial if some or most of the variables can be controlled in the data acquisition process, so as to generate enough data for the required estimations in limited recording time.
Further, the context for which the causal relation was designed influences the data collection process in terms of the location, other traffic participants and the general setting.
Due to the robust nature of causal theory, it is less important that the frequency of occurrence of each parameter combination is met exactly as in reality, although it remains necessary to adhere to the defined context.
Nevertheless, for accurate results it remains crucial to obtain enough data points for the aforementioned relevant variables to sufficiently estimate the CPDs, as marginalization according to \eqref{eq:causal_effect_adjustmentset} strongly depends on accurately derived CPDs.
For example, collecting data using camera drones observing traffic might have very few data points including rain, heavy wind or gusts and may thus lead to issues once these weather-related variables become important to determine certain causal effects.
Exemplary requirements on the real world data acquisition of an adjustment set for the causal model of the coefficient of friction are presented in \autoref{tab:adjustment_cof}.

\begin{table}[htb!]
	\footnotesize
	\caption{Measurability of the CP \textsc{reduced coefficient of friction} and a criticality metric $\agg(\stndt,\btndt)$ based on the causal structure in \autoref{fig:causal_relation_cof}.}
	\begin{tabularx}{\linewidth}{p{3.9cm}X}
		\toprule
		{\bf Exposure and outcome} & {\bf Measurability} \\
		Coefficient of friction & Can be approximated from values provided by vehicle sensors.\\
		Aggregate of $\btn$ and $\stn$ & Computed based on $\btn$ and $\stn$. \\
		Break threat number & Computed from the required and the minimal available longitudinal acceleration. \\
		Steer threat number & Computed from the required and the minimal available lateral acceleration. \\
		Required longitudinal acceleration & Computed based on ego vehicle forward velocity, planned acceleration and planned steering. \\
		Required lateral acceleration & Computed based on planned acceleration and planned steering. \\
		Maximal available longitudinal acceleration & Computed based on the ego vehicle longitudinal acceleration provided by the physical model, the ego vehicle tire type, the ego vehicle tire wetgrip and the ego vehicles maximal braking torque. \\
		Maximal available lateral acceleration & Computed based on the ego vehicle lateral acceleration provided by the physical model. \\\bottomrule
	\end{tabularx}
\end{table}

\subsection{Simulation Environments}
During the modeling and refinement of the causal relation, variables may occur that are not yet represented in the simulation environment (e.g.\ black ice) or known to be ignored in the computation of variables that are descendants with respect to the causal structure (e.g.\ black ice in the computation of the road-tire friction).
In these cases, implementing a causal relation in a simulation leads to requirements on the simulator itself, since the edges of the causal structure provide important information about the relevant interfaces of every simulation model, that can be seen as a necessary condition for the model to be valid.
Further, once a causal relation has been plausibilized and shall be used inside the simulation to demonstrate the effectiveness of a given SP, the validity of the simulation environment regarding the context of the causal relation is essential.
As causal theory provides us with a formula for the causal effect that is independent of the choice of the adjustment set, we may also derive requirements regarding the validity of the simulation from this, as the  same must be true for the causal effect in the simulation.
If this independence is not given, validity issues are likely to exist within the simulation environment. 
Whether the presented causal theory is able to constructively determine the underlying problem behind such validity issues is an open question and may be considered future work.
Once the validity of the simulation in the given context is assured, we further require the simulation to be able to record and even control every variable in the adjustment set.

\section{Conclusion and Future Work}
\label{sec:conclusion}

In this work, we laid the ground work for the application of causal theory to the problem of ensuring the safe operation of ADSs. In particular, we 
introduced a formalization of causal queries that can be asked during a criticality analysis as to gain a causal understanding  of 
criticality phenomena. For this, we defined a causal relation as a causal structure with focus on modeling the effect of a CP on criticality, as measured by a suitable metric, together with a context linked to a domain ontology. Moreover, we took first steps towards the assessment of a causal relation's modeling quality by introducing various comparative quantities. 
Our main example, for which a quantification is future work, is a causal relation for the CP \textsc{reduced coefficient of friction} and its effect on the ADS-operated vehicle's acceleration required by the driving task.  

This article leaves open possible extensions of the presented framework and its application. We now briefly elaborate on topics for future work and their utilization in the automotive domain.

{\it Temporal Causal Dependencies:} 
The causal relation of \autoref{sec:example} only features an ego vehicle. 
In general, however, the mutual interactions between traffic participants are integral to traffic but can also introduce cycles in the graph when considered on scenario level.
For this, we propose the usage of dynamic causal models \cite{blondel2017, srinivasan2021}, where each node represents a stochastic process and edges are modeled with consideration of time, allowing for mutual interactions without introducing cycles.

{\it Modularity of Causal Models (regarding Clustered Components):} 
Due to the combinatorial character of contexts, and thus the deduced causal models, regarding various traffic participants, it is beneficial to the modeling process to simply be add or remove entities from causal models.
Additionally, causal models need to be updateable to a changing traffic reality as e.g. demonstrated by the introduction of personal mobility devices.
Due to the similarities between causal models and fault trees, methods working towards the modularization of fault trees and already known issues thereof should be considered when assessing the modularization of causal models.
A similar approach to modularization for Bayesian networks has been derived by Koller and Pfeffer \cite{koller2013}. 

{\it Modularity of Causal Models (regarding Criticality Phenomena):}
So far we have only considered causal models for single criticality phenomena.
However, synergies between criticality phenomena must be examined to obtain a complete understanding of the phenomenon in each context.
Since there is a large variety of criticality phenomena, we propose the examination of interlocking of causal models for different criticality phenomena as future work.
One could imagine that the causal models might be stored in a connected database, thus sharing knowledge of the possibly relevant synergies, similar to the CHIELD database for causal hypotheses in evolutionary linguistics \cite{chield2020}.

{\it Plausibilization of Causal Models:}
The presented method focuses on theoretical foundations of the analysis of causal models.
Thus, as a next step towards the derivation of SPs, it is necessary to derive realistic distributions for the adjustment set variables, the CP and the criticality metric.
Since the results provided by the causal analysis may only be as good as the measured data, it is imperative to define guidelines under which the quantitative analysis provides significant results.

{\it Calculus for Stochastic Interventions:}
For now, we have only considered interventions that set variables to a specific value.
However, stochastic interventions \cite{Correa2020} provide a theory to adjust the distributions of the variables of a causal model and as such might provide useful information about the model or can be used to model SPs as the change of the distribution in a variable.

{\it Overall Safety Argumentation:}
The 2017 report of the German ethics commission on the topic of automated and connected driving requires the demonstrated reduction of harm to be causal in nature: 'The licensing of automated systems is not justifiable unless it promises to produce at least a diminution in harm compared with human driving' \cite[p. 10]{fabio2017ethics}. 
For this, the causal models presented in this work might be applied in a more abstract overarching safety argumentation.
Particularly, the fundamental question whether the introduction of an ADS results in less harm than had it not been introduced directly leads to a causative understanding of a positive risk balance.

\bibliographystyle{ieeetr}
\bibliography{Literature.bib}

\begin{thebibliography}{10}

\bibitem{sae2021definitions}
{SAE International}, ``{J3016: Taxonomy and Definitions for Terms Related to
  Driving Automation Systems for On-Road Motor Vehicles},'' 2021.

\bibitem{neucrit21}
C.~Neurohr, L.~Westhofen, M.~Butz, M.~H. Bollmann, U.~Eberle, and R.~Galbas,
  ``{Criticality Analysis for the Verification and Validation of Automated
  Vehicles},'' {\em IEEE Access}, vol.~9, pp.~18016--18041, 2021.

\bibitem{pearl2009causal}
J.~Pearl {\em et~al.}, ``{Causal inference in statistics: An overview},'' {\em
  Statistics surveys}, vol.~3, pp.~96--146, 2009.

\bibitem{poddey2019validation}
A.~Poddey, T.~Brade, J.~E. Stellet, and W.~Branz, ``{On the Validation of
  Complex Systems Operating in Open Contexts},'' {\em arXiv preprint
  arXiv:1902.10517}, 2019.

\bibitem{neurfund20}
C.~Neurohr, L.~Westhofen, T.~Henning, T.~de~Graaff, E.~Möhlmann, and E.~Böde,
  ``{Fundamental Considerations around Scenario-Based Testing for Automated
  Driving},'' in {\em 2020 IEEE Intelligent Vehicles Symposium (IV)},
  pp.~121--127, 2020.

\bibitem{iso26262}
{International Organization for Standardization}, ``{ISO 26262: Road vehicles
  -- Functional safety},'' {2018}.

\bibitem{westhofen2022metrics}
L.~Westhofen, C.~Neurohr, T.~Koopmann, M.~Butz, B.~Schütt, F.~Utesch,
  B.~Neurohr, C.~Gutenkunst, and E.~Böde, ``Criticality {Metrics} for
  {Automated} {Driving}: {A} {Review} and {Suitability} {Analysis} of the
  {State} of the {Art},'' {\em Archives of Computational Methods in
  Engineering}, 2022.

\bibitem{pearl_2009}
J.~Pearl, {\em Causality}.
\newblock Cambridge University Press, 2~ed., 2009.

\bibitem{Tian1998}
J.~Tian, A.~Paz, and J.~Pearl, ``{Finding Minimal D-separators},'' tech. rep.,
  University of California, Los Angeles, CA, 1998.

\bibitem{ankan_pgmpy_2015}
A.~Ankan and A.~Panda, ``pgmpy: {Probabilistic} {Graphical} {Models} using
  {Python},'' in {\em Proceedings of the 14th {Python} in {Science}
  {Conference}} (K.~Huff and J.~Bergstra, eds.), pp.~6 -- 11, 2015.

\bibitem{textor_2011_DAGitty}
J.~Textor, J.~Hardt, and S.~Knüppel, ``{DAGitty: A Graphical Tool for
  Analyzing Causal Diagrams},'' {\em Epidemiology (Cambridge, Mass.)}, vol.~22,
  p.~745, 2011.

\bibitem{sharma2020}
A.~Sharma and E.~Kiciman, ``Dowhy: An end-to-end library for causal
  inference,'' {\em arXiv preprint arXiv:2011.04216}, 2020.

\bibitem{peters2017}
J.~Peters, D.~Janzing, and B.~Schölkopf, {\em {Elements of Causal Inference:
  Foundations and Learning Algorithms}}.
\newblock The MIT Press, 2017.

\bibitem{hallerbach2022simulation}
S.~Hallerbach, U.~Eberle, and F.~Köster, {\em {Simulation-Enabled Methods for
  Development, Testing, and Validation of Cooperative and Automated Vehicles}},
  pp.~30--41.
\newblock Zenodo, 2022.

\bibitem{marcot2019advances}
B.~G. Marcot and T.~D. Penman, ``{Advances in Bayesian network modelling:
  Integration of modelling technologies},'' {\em Environmental modelling \&
  software}, vol.~111, pp.~386--393, 2019.

\bibitem{weber2012overview}
P.~Weber, G.~Medina-Oliva, C.~Simon, and B.~Iung, ``Overview on bayesian
  networks applications for dependability, risk analysis and maintenance
  areas,'' {\em Engineering Applications of Artificial Intelligence}, vol.~25,
  no.~4, pp.~671--682, 2012.

\bibitem{rauschenbach2017probabilistische}
M.~Rauschenbach, {\em {Probabilistische Grundlage zur Darstellung integraler
  Mehrzustands-Fehlermodelle komplexer technischer Systeme}}.
\newblock PhD thesis, 2017.

\bibitem{rauschenbach2019quantitative}
M.~Rauschenbach and J.~Nuffer, ``{Quantitative FMEA and Functional Safety
  Metrics Evaluation in Bayesian Networks},'' pp.~2475--2482, 2019.

\bibitem{wheeler2015initial}
T.~A. Wheeler, M.~J. Kochenderfer, and P.~Robbel, ``Initial scene
  configurations for highway traffic propagation,'' in {\em 2015 IEEE 18th
  International Conference on Intelligent Transportation Systems},
  pp.~279--284, 2015.

\bibitem{maier_bayesiansafety_2022}
R.~Maier and J.~Mottok, ``{BayesianSafety} - {An} {Open}-{Source} {Package} for
  {Causality}-{Guided}, {Multi}-model {Safety} {Analysis},'' in {\em Computer
  {Safety}, {Reliability}, and {Security}} (M.~Trapp, F.~Saglietti,
  M.~Spisländer, and F.~Bitsch, eds.), (Cham), pp.~17--30, Springer
  International Publishing, 2022.

\bibitem{iso21448}
{International Organization for Standardization}, ``{ISO 21448: Road vehicles
  -- Safety of the intended functionality},'' {2022}.

\bibitem{vesely1981fault}
W.~E. Vesely, F.~F. Goldberg, N.~H. Roberts, and D.~F. Haasl, ``{Fault Tree
  Handbook},'' tech. rep., Nuclear Regulatory Commission Washington DC, 1981.

\bibitem{ishimatsu2010modeling}
T.~Ishimatsu, N.~G. Leveson, J.~Thomas, M.~Katahira, Y.~Miyamoto, and H.~Nakao,
  ``{Modeling and Hazard Analysis using STPA},'' 2010.

\bibitem{kramer2020identification}
B.~Kramer, C.~Neurohr, M.~B{\"u}ker, E.~B{\"o}de, M.~Fr{\"a}nzle, and W.~Damm,
  ``{Identification and Quantification of Hazardous Scenarios for Automated
  Driving},'' in {\em Model-Based Safety and Assessment} (M.~Zeller and
  K.~H{\"o}fig, eds.), (Cham), pp.~163--178, Springer International Publishing,
  2020.

\bibitem{bode2019identifikation}
E.~B{\"o}de, M.~B{\"u}ker, W.~Damm, M.~Fr{\"a}nzle, B.~Kramer, C.~Neurohr, and
  S.~Vander~Maelen, ``{Identifikation und Quantifizierung von
  Automationsrisiken f{\"u}r hochautomatisierte Fahrfunktionen},'' tech. rep.,
  OFFIS e.V., 2019.

\bibitem{beck2022phenomenon}
H.~N. Beck, N.~F. Salem, V.~Haber, M.~Rauschenbach, and J.~Reich,
  ``{Phenomenon-Signal Model: Formalisation, Graph and Application},'' {\em
  arXiv preprint arXiv:2207.09996}, 2022.

\bibitem{linnhoff2021}
C.~Linnhoff, P.~Rosenberger, S.~Schmidt, L.~Elster, R.~Stark, and H.~Winner,
  ``{Towards Serious Perception Sensor Simulation for Safety Validation of
  Automated Driving - A Collaborative Method to Specify Sensor Models},'' in
  {\em 24th International Conference on Intelligent Transportation Systems
  (ITSC)}, (Darmstadt), IEEE, 2021.
\newblock Veranstaltungstitel: 24th International Conference on Intelligent
  Transportation Systems (ITSC).

\bibitem{linnhoff2022}
C.~Linnhoff, K.~Hofrichter, L.~Elster, P.~Rosenberger, and H.~Winner,
  ``Measuring the influence of environmental conditions on automotive lidar
  sensors,'' {\em Sensors}, vol.~22, no.~14, 2022.

\bibitem{westhofen2022ontologies}
L.~Westhofen, C.~Neurohr, M.~Butz, M.~Scholtes, and M.~Schuldes, ``{Using
  Ontologies for the Formalization and Recognition of Criticality for Automated
  Driving},'' {\em IEEE Open Journal of Intelligent Transportation Systems},
  vol.~3, pp.~519--538, 2022.

\bibitem{asamOpenXOntology}
{A}ssociation for {S}tandardization~of {A}utomation and {M}easuring~{S}ystems
  (ASAM), ``{ASAM OpenXOntology},'' 2022.

\bibitem{scholtes2021layer}
M.~Scholtes, L.~Westhofen, L.~R. Turner, K.~Lotto, M.~Schuldes, H.~Weber,
  N.~Wagener, C.~Neurohr, M.~H. Bollmann, F.~Körtke, J.~Hiller, M.~Hoss,
  J.~Bock, and L.~Eckstein, ``{6-Layer Model for a Structured Description and
  Categorization of Urban Traffic and Environment},'' {\em IEEE Access},
  vol.~9, pp.~59131--59147, 2021.

\bibitem{mackay2003}
D.~J.~C. MacKay, {\em {Information Theory, Inference and Learning Algorithms}}.
\newblock Cambridge University Press, 2003.

\bibitem{janzing2013}
D.~Janzing, D.~Balduzzi, M.~Grosse-Wentrup, and B.~Schölkopf, ``{Quantifying
  causal influences},'' {\em The Annals of Statistics}, vol.~41, no.~5,
  pp.~2324 -- 2358, 2013.

\bibitem{gretton2012}
A.~Gretton, K.~M. Borgwardt, M.~J. Rasch, B.~Sch{\"o}lkopf, and A.~Smola, ``A
  kernel two-sample test,'' {\em Journal of Machine Learning Research},
  vol.~13, 2012.

\bibitem{wronski2014}
L.~Wronski, {\em {Reichenbach’s Paradise Constructing the Realm of
  Probabilistic Common “Causes”}}.
\newblock De Gruyter, 2014.

\bibitem{Persson2006}
B.~N.~J. Persson, ``Rubber friction: role of the flash temperature.,'' {\em
  Journal of physics. Condensed matter : an Institute of Physics journal},
  vol.~18 32, pp.~7789--823, 2006.

\bibitem{Jansson2005}
J.~Jansson, {\em {Collision Avoidance Theory with Application to Automotive
  Collision Mitigation}}.
\newblock PhD thesis, Linköping University, 2005.

\bibitem{brach2011}
R.~Brach and M.~Brach, ``{The Tire-Force Ellipse (Friction Ellipse) and Tire
  Characteristics},'' {\em SAE 2011 World Congress and Exhibition}, vol.~2296,
  2011.

\bibitem{riedmaier2020}
S.~Riedmaier, D.~Schneider, D.~Watzenig, F.~Diermeyer, and B.~Schick, ``{Model
  Validation and Scenario Selection for Virtual-Based Homologation of Automated
  Vehicles},'' {\em Applied Sciences}, vol.~11, 2020.

\bibitem{rosenberger2019towards}
P.~Rosenberger, J.~T. Wendler, M.~F. Holder, C.~Linnhoff, M.~Bergh{\"o}fer,
  H.~Winner, and M.~Maurer, ``{Towards a Generally Accepted Validation
  Methodology for Sensor Models - Challenges, Metrics, and First Results},''
  2019.

\bibitem{rosenberger2020sequential}
P.~Rosenberger, M.~F. Holder, N.~Cianciaruso, P.~Aust, J.~F. Tamm-Morschel,
  C.~Linnhoff, and H.~Winner, ``{Sequential lidar sensor system simulation: a
  modular approach for simulation-based safety validation of automated
  driving},'' {\em Automotive and Engine Technology}, vol.~5, no.~3,
  pp.~187--197, 2020.

\bibitem{blondel2017}
G.~Blondel, M.~Arias, and R.~Gavaldà, ``{Identifiability and Transportability
  in Dynamic Causal Networks},'' {\em International Journal of Data Science and
  Analytics}, vol.~3, 2017.

\bibitem{srinivasan2021}
R.~Srinivasan, J.~J.~R. Lee, R.~Bhattacharya, and I.~Shpitser, ``Path dependent
  structural equation models,'' in {\em Proceedings of the Thirty-Seventh
  Conference on Uncertainty in Artificial Intelligence} (C.~de~Campos and M.~H.
  Maathuis, eds.), vol.~161 of {\em Proceedings of Machine Learning Research},
  pp.~161--171, PMLR, 2021.

\bibitem{koller2013}
D.~Koller and A.~Pfeffer, ``{Object-Oriented Bayesian Networks},'' 2013.

\bibitem{chield2020}
S.~G. Roberts, A.~Killin, A.~Deb, C.~Sheard, S.~J. Greenhill, K.~Sinnem\"{a}ki,
  J.~Segovia-Mart\'{i}n, J.~N\"{o}lle, A.~Berdicevskis, A.~Humphreys-Balkwill,
  H.~Little, C.~Opie, G.~Jacques, L.~Bromham, P.~Tinits, R.~M. Ross, S.~Lee,
  E.~Gasser, J.~Calladine, M.~Spike, S.~F. Mann, O.~Shcherbakova, R.~Singer,
  S.~Zhang, A.~Ben\'{i}tez-Burraco, C.~Kliesch, E.~Thomas-Colquhoun,
  H.~Skirg\r{a}rd, M.~Tamariz, S.~Passmore, T.~Pellard, and F.~Jordan,
  ``{CHIELD: the causal hypotheses in evolutionary linguistics database},''
  {\em Journal of Language Evolution}, 2020.

\bibitem{Correa2020}
J.~Correa and E.~Bareinboim, ``{A Calculus for Stochastic Interventions: Causal
  Effect Identification and Surrogate Experiments},'' {\em Proceedings of the
  AAAI Conference on Artificial Intelligence}, vol.~34, no.~06,
  pp.~10093--10100, 2020.

\bibitem{fabio2017ethics}
U.~Di~Fabio, M.~Broy, R.~Br{\"u}ngger, U.~Eichhorn, A.~Grunwald, D.~Heckmann,
  E.~Hilgendorf, H.~Kagermann, A.~Losinger, M.~Lutz-Bachmann, {\em et~al.},
  ``{Ethics Commission: Automated and Connected Driving},'' tech. rep., German
  Federal Ministry of Transport and Digital Infrastructure, 2017.

\bibitem{khaleghian2017}
S.~Khaleghian, A.~Emami, and S.~Taheri, ``A technical survey on tire-road
  friction estimation,'' {\em Friction}, vol.~5, pp.~123--146, 2017.

\end{thebibliography}

\begin{acks}
	The research leading to these results is funded by the German Federal Ministry for Economic Affairs and Climate Action within the project 'VVM -- Verification \& Validation Methods for Automated Vehicles Level 4 and 5'.
\end{acks}

\appendix
\section{Appendix}
\label{sec:appendix}

\subsection{Detailed List of the Variables for the Causal Structure in \autoref{fig:causal_relation_cof}}
\label{subsec:appendix_1}

{\footnotesize
\begin{longtable}{>{\raggedright\arraybackslash}p{2.85cm}>{\raggedright\arraybackslash}p{7.33cm}>{\raggedright\arraybackslash}p{2.66cm}}
	\caption{Properties of the variables in the example causal relation for \textsc{Reduced Coefficient of Friction}.} \\
	\toprule
	{\bf Variable} & {\bf Interpretation} & {\bf Range and Unit} \\\endhead
	\bottomrule\endfoot
	Date and time of day & Time in a fixed time zone &  $\{0,$\dots$,23\} \times \{0,$\dots$,59\}$ \\
	Global location & Area of the world where the scenario takes place &  $\left[-\frac{\pi}{2}, \frac{\pi}{2}\right] \times \left[-\pi, \pi\right]$ \\
	Weather & Weather at the location where the scenario takes place & \{warm, cold, \dots\} $\times$ \{clear sky, cloudy, \dots\} \\
	Humidity & Average current relative humidity over the area of the scenario &  [0, $\infty$), \si{\percent} \\
	Dew point & Average current dew point over the area of the scenario & $[0, \infty)$, \si{\kelvin} \\
	Precipitation & Average precipitation in the general area of the scenario &  $[0, \infty)$, \si{\milli\meter\per\square\meter\per\hour} \\
	Road list & List of relevant road networks & \{road\_1, \dots, road\_n\} \\
	Relative position of ego & Position of ego relative to fixed origin &  $(-\infty, \infty)^3$, \si{\meter} \\
	Speed limit of ego & Minimal speed limit for ego for the next 200m &  $[0, \infty)$, \si{\meter\per\second} \\
	Ego distance to next ISRL & Distance of ego to next intersection stopping requirement location, e.g.\ traffic light or stop sign  & $[0, \infty)$, \si{\meter} \\
	Target position of ego & Planned target location of ego relative to the origin &  $(-\infty,\infty)^3$, \si{\meter} \\
	Ego road curvature & Curvature of the road segment the ego is currently driving on &  $[0,\infty)$, \si{\per\meter} \\
	Global illumination & Average illuminance in a designated area around the ego & $[0, \infty)$, \si{\candela\per\square\meter} \\
	Air temperature & Average air temperature throughout the scenario state space & $[0, \infty)$, \si{\kelvin} \\
	Degree of Wetness & Average thickness of water film on the road in area around ego &  $[0, \infty)$, \si{\meter} \\
	Winter slipperiness & Relative area of the road covered with ice in area around ego &  $[0, 100]$, \si{\percent} \\
	Road surface material & Material of the road surface in a designated area around ego &  \{asphalt, gravel, \dots\} \\
	Road surface contamination & Average contamination of the road surface, e.g. leaves or oil & $[0, \infty)^n$ for $n = |$ \{leaves, soil, oil, \dots\} $|$ \\
	Road surface degradation & General condition of the road surface using the pavement condition index & \{0, \dots, 100\} \\
	Road surface roughness & Aggregated distance of sample points to the mean road surface &  $[0, \infty)$, \si{\meter} \\
	Planned steering & Planned average steering of ego for the next $x$ seconds &  $[-\frac{\pi}{2}, \frac{\pi}{2}]$ \\
	Planned acceleration & Planned average acceleration of ego for the next $x$ seconds &  $[0, \infty)$, \si{\meter\per\second\squared} \\
	Ego tire temperature & Temperature of the tires & $[0, \infty)$, \si{\kelvin} \\
	Ego tire flash temp. & Aggregate of the temperature of the tires at the contact patch & $[0, \infty)$, \si{\kelvin} \\
	Tire type & General classification of the tire type & \{Touring, Winter, \dots\} \\
	Wet grip & EU tire label regarding wetgrip & \{A,B,C,E,F\} \\
	Tire pressure & Aggregated tire pressure, e.g. minimum or average & $[0, \infty)$, \si{\newton\per\square\meter} \\
	Maximal braking torque & Ability to brake at the brakes themselves &  $[0, \infty)$, \si{\newton\meter} \\
	Ego vehicle mass & Mass of ego including vehicle load &  $(0, \infty)$, \si{\kilogram} \\
	Forward velocity of ego & Velocity of the ego vehicle relative to the road &  $\mathbb{R}^3$, \si{\meter\per\second} \\
	Hub velocity of ego & Velocity of the ego measured by the system &  $\mathbb{R}^3$, \si{\meter\per\second} \\
	Coefficient of friction & Average coefficient of friction between road surface and tires \cite{khaleghian2017} &  $[0, \infty), 1$ \\
	Max. avail. long. dec. & Maximal available longitudinal deceleration of the ego vehicle & $[0, \infty)$, \si{\meter\per\square\second} \\
	Max. avail. lat. dec. & Maximal available lateral deceleration of the ego vehicle &  $[0, \infty)$, \si{\meter\per\square\second} \\
	Max. req. long. dec. & Maximal required longitudinal deceleration from the driving task & $[0, \infty)$, \si{\meter\per\square\second} \\
	Max. req. lat. dec. & Maximal required lateral deceleration from the driving task &  $[0, \infty)$, \si{\meter\per\square\second} \\
	$\stndt$ & Steer thread number from the driving task & $[0, \infty)$, $1$ \\
	$\btndt$ & Brake thread number from the driving task & $[0, \infty)$, $1$ \\
	agg($\stndt$, $\btndt$) & Aggregate of $\btndt$ and $\stndt$ & $[0, \infty)$, $1$
\end{longtable}
}

\end{document}